%% file: icml2025.tex
\documentclass{article}

\usepackage{microtype}
\usepackage{booktabs} 
\usepackage{graphicx}
\usepackage{subcaption} 
\usepackage{wrapfig}
\usepackage{hyperref}

\usepackage[accepted]{icml2025} 

\makeatletter
\renewcommand{\Notice@String}{}
\makeatother

\usepackage{amsmath}
\usepackage{amssymb}
\usepackage{mathtools}
\usepackage{amsthm}
\usepackage{wrapfig}

\usepackage{xcolor}
\usepackage{multirow}
\usepackage[capitalize,noabbrev]{cleveref}

\theoremstyle{plain}

\theoremstyle{definition}

\theoremstyle{remark}

\usepackage[textsize=tiny]{todonotes}
\usepackage{algorithm}
\usepackage{algpseudocode}
\usepackage{amsmath}

\icmltitlerunning{Enhancing Diffusion Posterior Sampling
for Inverse Problems by Integrating Crafted Measurements}

\begin{document}

\twocolumn[
\icmltitle{Enhancing Diffusion Posterior Sampling \\for Inverse Problems by Integrating Crafted Measurements}



\icmlsetsymbol{corre}{*}

\begin{icmlauthorlist}
\icmlauthor{Shijie Zhou}{ub}
\icmlauthor{Huaisheng Zhu}{psu}
\icmlauthor{Rohan Sharma}{ub}
\icmlauthor{Jiayi Chen}{fju} 
\icmlauthor{Ruiyi Zhang}{adobe}
\icmlauthor{Kaiyi Ji}{ub}
\icmlauthor{Changyou Chen}{ub}
\end{icmlauthorlist}
\icmlcorrespondingauthor{Shijie Zhou}{shijiezh@buffalo.edu}
\icmlcorrespondingauthor{Changyou Chen}{changyou@buffalo.edu}
\icmlaffiliation{ub}{University at Buffalo}
\icmlaffiliation{adobe}{Adobe Research}
\icmlaffiliation{psu}{The Pennsylvania State University}
\icmlaffiliation{fju}{Fujian Normal University}



\icmlkeywords{Machine Learning, ICML}

\vskip 0.3in
]



\printAffiliationsAndNotice{} 

\input{main/abstract}
\input{main/intro}
\input{main/background}
\input{main/method}
\input{main/experiments}
\input{main/conlusion}

\bibliography{icml2025}
\bibliographystyle{icml2025}

\input{main/appendix}



\end{document}

%% file: main/abstract.tex
\begin{abstract}
Diffusion models have emerged as a powerful foundation model for visual generations. With an appropriate sampling process, it can effectively serve as a generative prior for solving general inverse problems. Current posterior sampling-based methods take the measurement (i.e., degraded image sample) into the posterior sampling to infer the distribution of the target data (i.e., clean image sample). However, in this manner, we show that high-frequency information can be prematurely introduced during the early stages, which could induce larger posterior estimate errors during restoration sampling. To address this observation, we first reveal that forming the log-posterior gradient with the noisy measurement ( i.e., noisy measurement from a diffusion forward process) instead of the clean one can benefit the early posterior sampling. Consequently, we propose a novel diffusion posterior sampling method DPS-CM, which incorporates a Crafted Measurement (i.e., noisy measurement crafted by a reverse denoising process, rather than constructed from the diffusion forward process) to form the posterior estimate. This integration aims to mitigate the misalignment with the diffusion prior caused by cumulative posterior estimate errors. Experimental results demonstrate that our approach significantly improves the overall capacity to solve general and noisy inverse problems, such as Gaussian deblurring, super-resolution, inpainting, nonlinear deblurring, and tasks with Poisson noise, relative to existing approaches. Code is available at \url{https://github.com/sjz5202/DPS-CM}.
\end{abstract}

%% file: main/intro.tex
\section{Introduction}
Diffusion models \citep{ho2020denoising} have achieved remarkable generative performance on images \citep{amit2021segdiff, baranchuk2021label,brempong2022denoising}, videos \citep{singer2022make,wu2023tune}, audios \citep{popov2021grad,yang2023diffsound}, natural language \citep{austin2021structured, hoogeboom2021argmax, li2022diffusion} and molecular generation \citep{hoogeboom2022equivariant,jing2022torsional}. Besides its strong modeling capacity for complex and high dimensional data, diffusion models have exhibited a strong generative prior 
to form the diffusion conditional sampling 
\citep{song2020score} that can be harnessed for diffusion posterior sampling. In the context of noisy inverse problems, this sampling process effectively approximates precise data distributions from noisy and degraded measurements. Noisy inverse problems, such as super-resolution, inpainting, linear and nonlinear deblurring, are targeted to restore an unknown image $\boldsymbol{x}$ from its noise-corrupted measurement $\boldsymbol{y}$ given the corresponding forward measurement operators $\mathcal{A}(\cdot): \mathbb{R}^n \rightarrow \mathbb{R}^m$. Recently, Diffusion models have been extensively utilized for these tasks \citep{zhu2023denoising,song2022pseudoinverse,wang2022zero}, offering a robust framework for reconstructing high-quality images from degraded measurements. 

Current diffusion-based methods generally employ two distinct strategies to solve inverse problems. The first strategy is to train problem-specialized diffusion models \citep{saharia2022image,whang2022deblurring,luo2023refusion,chan2023sud2supervisiondenoisingdiffusion} given measurements and clean image pairs. In contrast, methods of the second strategy only capitalize on problem-agnostic pre-trained diffusion models to benefit the zero-shot diffusion restoration sampling by posterior estimate \citep{song2023solving,rout2024solving,peng2024improving,mardani2023variational} or enforcing data consistency \citep{chung2022improving}. In this paper, we concentrate on the second manner, the posterior estimate for sampling, to solve noisy inverse problems universally.

Posterior estimate based methods enable controllable generations \citep{dhariwal2021diffusion} via diffusion conditional reverse-time SDE \citep{song2020score}, such as classifier guidance \citep{dhariwal2021diffusion}, loss-guided diffusion \citep{song2023loss}. Applying Bayesian, the gradient of log posterior $\nabla_{\boldsymbol{x}_t} \log p_t\left(\boldsymbol{x}_t| \boldsymbol{y}\right)$ can be easily adopted into conditional reverse-time SDE sampling as a log-likelihood gradient term $\nabla_{\boldsymbol{x}_t} \log p\left(\boldsymbol{y} | \boldsymbol{x}_t\right)$ and an unconditional prior term $\nabla_{\boldsymbol{x}_t} \log p_t\left(\boldsymbol{x}_t\right)$. In the context of solving inverse problems, however, $ p\left(\boldsymbol{y} | \boldsymbol{x}_t\right)$ is an intractable distribution due to the unclear dependency between the measurement $\boldsymbol{y}$ and the diffusion generation $\boldsymbol{x}_t$ at time $t$. To tackle this issue, existing posterior estimate methods for inverse problems, such as Diffusion Posterior Sampling (DPS \citep{chung2022diffusion}), form the measurement model $p\left(\boldsymbol{y} | \hat{\boldsymbol{x}}_0\right)$ as the likelihood estimate, where $\hat{\boldsymbol{x}}_0$ is the denoising prediction given diffusion intermediate output $\boldsymbol{x}_t$. This process maps $\hat{\boldsymbol{x}}_0$ into the measurement $\boldsymbol{y}$'s space, which can be interpreted as a reconstruction loss guidance to push $\mathcal{A}\left(\hat{\boldsymbol{x}}_0\right)$ close to $\boldsymbol{y}$, while narrowing the gap between $\boldsymbol{x}_t$ and the clean image $\boldsymbol{x}$.

However, via empirical examples in Section~\ref{sec:3_1}, we show that solving inverse problems in the manner of diffusion posterior sampling follows a similar pattern of diffusion reverse process \citep{yang2023diffusion}, i.e., focusing on low-frequency recovery at first and posing increasing attention on high-frequency generation in the late stages.
With such observation, the log posterior gradient estimate $\nabla_{\boldsymbol{x}_t} \log p\left(\boldsymbol{y} | \hat{\boldsymbol{x}}_0\right)$ in DPS with sharp measurement $\boldsymbol{y}$ will easily introduce abrupt high-frequency gradient signals for the subsequent step, which unfits the appropriate input pattern for the pre-trained model $\boldsymbol{s}_\theta\left(\boldsymbol{x}_t, t\right)$ during the early stages with large $t$.  
In fact, \cite{song2023loss} also notes that DPS significantly miscalculates the scale of the guidance term with different variance levels which results in accumulated errors in posterior sampling. 
In Section~\ref{sec:3_1}, we have similar observations that DPS amplifies posterior sampling errors. We also find that applying the likelihood estimate $p\left(\boldsymbol{y}_t | \hat{\boldsymbol{x}}_0\right)$ with randomly sampled \textbf{noisy measurement} $\boldsymbol{y}_t$ instead of the clean measurement $\boldsymbol{y}$ in the DPS
for each timestep $t$ leads to a smaller approximation error during the early stages and thus benefits the restoration generation. Posterior approximation with \textbf{noisy measurement} $\boldsymbol{y}_t$ at timestep $t$, compared with the clean one, has the advantage that it adaptively matches the frequency pattern of the diffusion model's generation at the timestep $t$.

Therefore, with this insight, we propose the posterior approximation that leads to less high-frequency signal recovery during the early stages. Specifically, we propose the Diffusion Posterior Sampling with \textbf{Crafted Measurements} (\textbf{DPS-CM}), which can introduce a less biased posterior estimate by combining crafted measurement $\mathbf{y}_t$ \footnote{In this work, we denote the crafted measurement as $\mathbf{y}_t$, which is the intermediate generation of the diffusion reverse process with diffusion model $\theta$, and the noisy measurement as $\boldsymbol{y}_t$, which is generated by the diffusion forward process as in Eq. \ref{eq:formxt}.}, \textbf{the intermediate generation of another diffusion reverse-time trajectory $\{\mathbf{y}_t\}_{t=0}^T$} from posterior $p\left(\mathbf{y}_t | \boldsymbol{y}\right)$. 
As $\{\mathbf{y}_t\}_{t=0}^T$ shares a similar frequency distribution pattern with the target generation trajectory $\{\boldsymbol{x}_t\}_{t=0}^T$, i.e., low-frequency recovery at first, the approximated log-likelihood gradient $\nabla_{\boldsymbol{x}_t} \log p\left(\hat{\mathbf{y}}_0 | \hat{\boldsymbol{x}}_0\right)$ will bring in less high-frequency gradient signal and thus benefit the subsequent generations. Besides, leveraging \textbf{crafted measurements} $\mathbf{y}_t$ brings a lower bias compared with directly using randomly sampled \textbf{noisy measurement} $\boldsymbol{y}_t$.  
Our extensive experiments results on various noisy linear inverse problems, e.g. super-resolution, random masked/fixed box inpainting, Gaussian/Motion deblurring,  and nonlinear inverse problems such as nonlinear deblurring, demonstrate that the proposed DPS-CM significantly outperforms existing unsupervised methods on both FFHQ \citep{karras2019style} and ImageNet \citep{deng2009imagenet} datasets while keeping the algorithm simplicity.  

%% file: main/background.tex
\section{Background}
\subsection{Diffusion Models}
Diffusion models \citep{ho2020denoising, song2020score} comprise two processes: a forward noising process in which the noise is progressively injected into the sample and a reverse denoising process for generation. Specifically, the forward process of Denoising Diffusion Probabilistic Models (DDPM) \citep{ho2020denoising} can be formulated by the variance preserving stochastic differential equation (VP-SDE) \citep{song2020score}:
\begin{equation}
    d \boldsymbol{x}=-\frac{\beta_t}{2} \boldsymbol{x} d t+\sqrt{\beta_t} d \boldsymbol{w}, \quad t \in[0, T],
\label{eq:vp}
\end{equation}
where $\{\beta_t\}_{t=0}^T$ is the designed noise schedule which is monotonically increased and $\boldsymbol{w}$ is the standard Wiener process. To generate samples of targeted data distribution from Gaussian noise $\boldsymbol{x}_T \sim \mathcal{N} (\mathbf{0}, \boldsymbol{I})$, the corresponding reverse SDE of Eq. \ref{eq:vp} is formulated as:
\begin{equation}
    d \boldsymbol{x}=\left[-\frac{\beta_t}{2} \boldsymbol{x}-\beta_t \nabla_{\boldsymbol{x}_t} \log p\left(\boldsymbol{x}_t\right)\right] d t+\sqrt{\beta_t} d \bar{\boldsymbol{w}},
    \label{eq:uncon}
\end{equation}
where $d \bar{\boldsymbol{w}}$ is the reverse standard Wiener process and $\nabla_{\boldsymbol{x}_t} \log p\left(\boldsymbol{x}_t\right)$ is the score function of denoised sample at time $t$. Approximating this score function is the key to solving the reverse generative process, which can be done by training a parameterized model $\boldsymbol{s}_{\theta}\left(\boldsymbol{x}_t, t\right)$ by denoising score matching \citep{song2019generative} expected on samples from the forward diffusion process:
\begin{equation}
    \boldsymbol{x}_t=\sqrt{\bar{\alpha}_t} \boldsymbol{x}_0+\sqrt{1-\bar{\alpha}_t} \boldsymbol{z}, \quad \boldsymbol{z} \sim \mathcal{N}(\mathbf{0}, \boldsymbol{I}).
    \label{eq:formxt}
\end{equation}
With a trained score model $\boldsymbol{s}_{\theta}$, applying Tweedie’s formula \citep{efron2011tweedie,dunn2005series}, we can approximate the mean $\hat{\boldsymbol{x}}_0$ of the posterior $p\left(\boldsymbol{x}_0 | \boldsymbol{x}_t\right)$ for the above case (VP-SDE) as: 
\begin{equation}
    \hat{\boldsymbol{x}}_0 :=\mathbb{E}\left[\boldsymbol{x}_0 | \boldsymbol{x}_t\right] \simeq \frac{1}{\sqrt{\bar{\alpha}(t)}}\left(\boldsymbol{x}_t+(1-\bar{\alpha}(t)) \boldsymbol{s}_{\theta}\left(\boldsymbol{x}_t, t\right)\right).
    \label{eq:x0hat}
\end{equation}
\subsection{Solving Inverse Problems with Diffusion Models}
Given a clean image $\boldsymbol{x}_0$ and a forward measurement operator $\mathcal{A}(\cdot)$, the general form of the noisy inverse problem can be stated as: 
\[
\boldsymbol{y}=\mathcal{A}\left(\boldsymbol{x}_0\right)+\boldsymbol{\eta},
\]
where an i.i.d. noise $\boldsymbol{\eta}\sim \mathcal{N}\left(\mathbf{0}, \sigma^2 \boldsymbol{I}\right)$ is added in $\mathcal{A}\left(\boldsymbol{x}_0\right)$ for the case of Gaussian noise to formulate the noisy inverse problem. Given measurements $\boldsymbol{y} \in \mathbb{R}^m$ and a known forward measurement operator $\mathcal{A}(\cdot): \mathbb{R}^n \rightarrow \mathbb{R}^m$, the target is to restore the unknown signal $\boldsymbol{x}_0 \in \mathbb{R}^n$.
Replacing the unconditional score $\nabla_{\boldsymbol{x}_t} \log p\left(\boldsymbol{x}_t \right)$ in Eq. \ref{eq:uncon} with the conditional score $\nabla_{\boldsymbol{x}_t} \log p\left(\boldsymbol{x}_t | \boldsymbol{y}\right)$ paves the way for solving inverse problems using diffusion models. Applying the Bayesian rule on the gradient of the posterior $\nabla_{\boldsymbol{x}_t} \log p\left(\boldsymbol{x}_t | \boldsymbol{y}\right)$, the reverse-time SDE of the posterior sampling is given by:
\begin{equation}
\begin{aligned}
d\boldsymbol{x} = & \left[-\frac{\beta_t}{2}\,\boldsymbol{x} - \beta_t\,\nabla_{\boldsymbol{x}_t}(\log p(\boldsymbol{x}_t) + \log p(\boldsymbol{y}|\boldsymbol{x}_t))\right]dt \\
& +\sqrt{\beta_t}d\bar{\boldsymbol{w}}. \label{eq:con}
\end{aligned}
\end{equation}
However, the gradient of the log-likelihood $\nabla_{\boldsymbol{x}_t} \log p\left(\boldsymbol{y} | \boldsymbol{x}_t\right)$ in Eq.\ref{eq:con} is intractable given $\boldsymbol{y}=\mathcal{A}\left(\boldsymbol{x}_0\right)+\boldsymbol{\eta}$. DDRM \citep{kawar2022denoising} addresses this issue by forming the conditional sampling in the spectral space without dealing with $\nabla_{\boldsymbol{x}_t} \log p\left(\boldsymbol{y} | \boldsymbol{x}_t\right)$, but with the restriction to handle complex inverse problems due to the SVD computation efficiency. DPS \citep{chung2022diffusion} provides a universal posterior approximation approach to deal with the intractable likelihood term $\nabla_{\boldsymbol{x}_t} \log p\left(\boldsymbol{y} | \boldsymbol{x}_t\right)$. Given the mean estimate $\hat{\boldsymbol{x}}_0$ of posterior $\mathbb{E}\left[\boldsymbol{x}_0 | \boldsymbol{x}_t\right]$ by applying Eq.\ref{eq:x0hat}, DPS assumes:
\begin{equation}
\begin{aligned}
p\left(\boldsymbol{y} | \boldsymbol{x}_t\right) & = \mathbb{E}_{\boldsymbol{x}_0 \sim p\left(\boldsymbol{x}_0 | \boldsymbol{x}_t\right)}\left[p\left(\boldsymbol{y} | \boldsymbol{x}_0\right)\right]\\
& \simeq  p\left(\boldsymbol{y} | \hat{\boldsymbol{x}}_0\right).\\
\end{aligned}
\label{eq:dps}
\end{equation}
As the production, they form the gradient of the posterior as:
\begin{equation}
    \nabla_{\boldsymbol{x}_t} \log p_t\left(\boldsymbol{x}_t | \boldsymbol{y}\right) \simeq \boldsymbol{s}_{\theta^*}\left(\boldsymbol{x}_t, t\right)-\rho \nabla_{\boldsymbol{x}_t}\left\|\boldsymbol{y}-\mathcal{A}\left(\hat{\boldsymbol{x}}_0\right)\right\|_2^2.
    \label{eq:post}
\end{equation}
Although DPS provides a straightforward and task-independent approximation, many existing works, such as \cite{peng2024improving,song2023loss}, point out this posterior estimate is biased.

%% file: main/method.tex
\section{Method: Diffusion Posterior Sampling with Crafted Measurements}
We propose DPS-CM, which performs diffusion posterior sampling from $p\left(\boldsymbol{x}_t | \mathbf{y}_t\right)$ 
with \textbf{Crafted Measurement} $\mathbf{y}_t$ belonging to another diffusion reverse trajectory $\{\mathbf{y}_t\}_{t=0}^T$ instead of the vanilla input $\boldsymbol{y}$. Specifically, applying Eq.\ref{eq:x0hat} on $\boldsymbol{x}_t$ and $\mathbf{y}_t$, we incorporate the tractable $p\left(\hat{\mathbf{y}}_0 | \hat{\boldsymbol{x}}_0\right)$ as the likelihood approximation of $p\left(\mathbf{y}_t | \boldsymbol{x}_t\right)$ to enable the posterior sampling from $p\left(\boldsymbol{x}_t | \mathbf{y}_t\right)$.
DPS-CM can be interpreted as pushing the bond between two noisy approximations $\hat{\mathbf{y}}_0$ and $\hat{\boldsymbol{x}}_0$ as the measurement model $\hat{\mathbf{y}}_0=\mathcal{A}\left(\hat{\boldsymbol{x}}_0\right)$. While $\hat{\mathbf{y}}_0$ is progressively recovered to the input measurement $\boldsymbol{y}$ via its diffusion denoising process, $\hat{\boldsymbol{x}}_0$ is expected to develop into the target ground truth $\boldsymbol{x}$ gradually during the synchronous reverse process with the measurement bond $\hat{\mathbf{y}}_0=\mathcal{A}\left(\hat{\boldsymbol{x}}_0\right)$. 
{\bf Roadmap.} In \ref{sec:3_1}, we explore the advantage of posterior sampling utilizing noisy measurements $\boldsymbol{y}_t$ sampled from the diffusion forward process over using the clean input $y$ in reducing intermediate generation errors in diffusion posterior sampling. In \ref{sec:3_2}, we formulate the DPS-CM in detail. 

\subsection{DPS Amplify Posterior Sampling Errors}\label{sec:3_1}

\begin{figure*}[ht!]
    \centering
    \begin{subfigure}[t]{0.32\textwidth}
        \includegraphics[width=\linewidth]{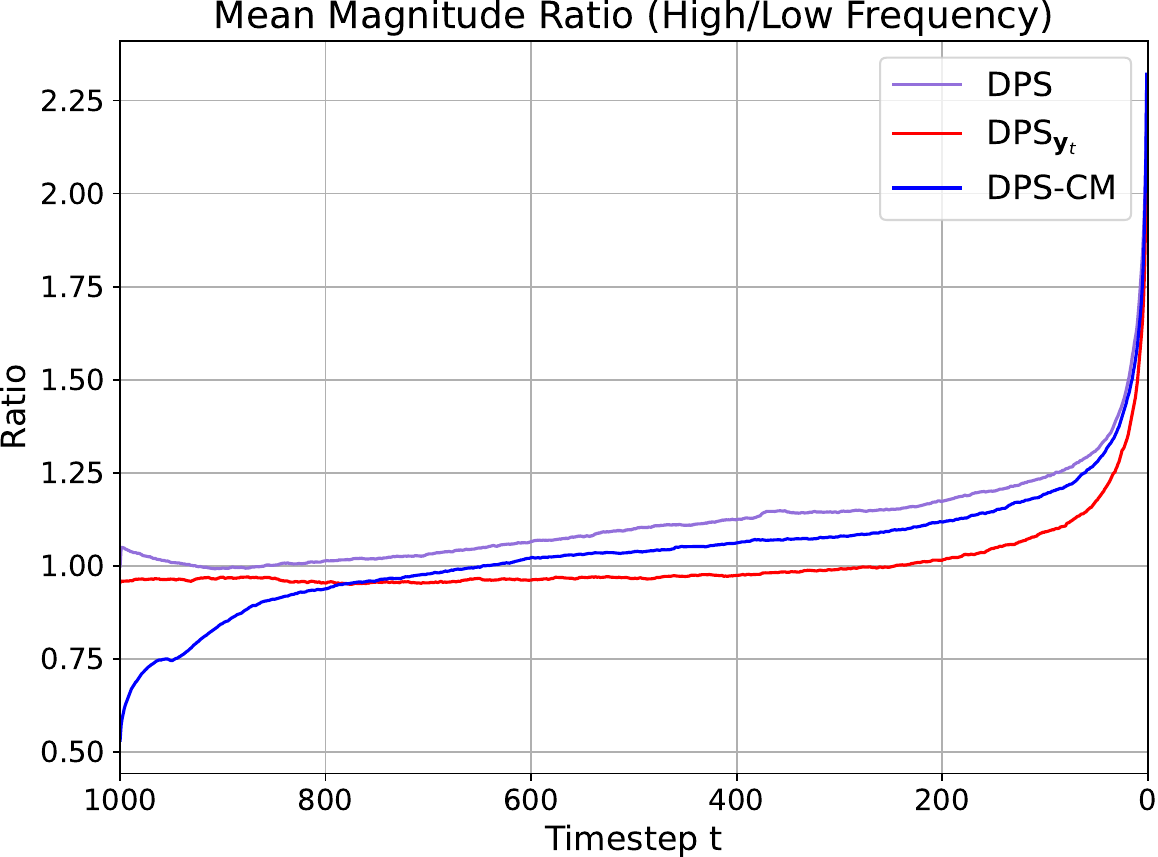}
        \caption{}
        \label{fig:freq}
    \end{subfigure}
    \hfill
    \begin{subfigure}[t]{0.32\textwidth}
        \includegraphics[width=\linewidth]{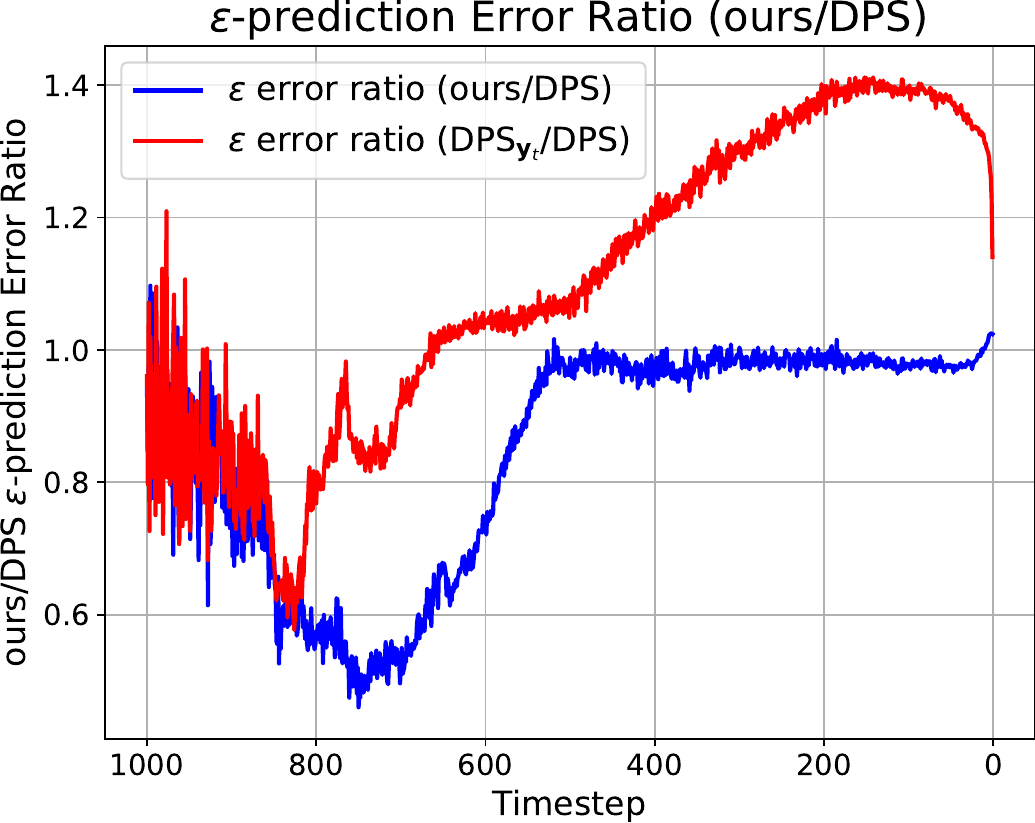}
        \caption{}
        \label{fig:epsilon}
    \end{subfigure}
    \hfill
    \begin{subfigure}[t]{0.32\textwidth}
        \includegraphics[width=\linewidth]{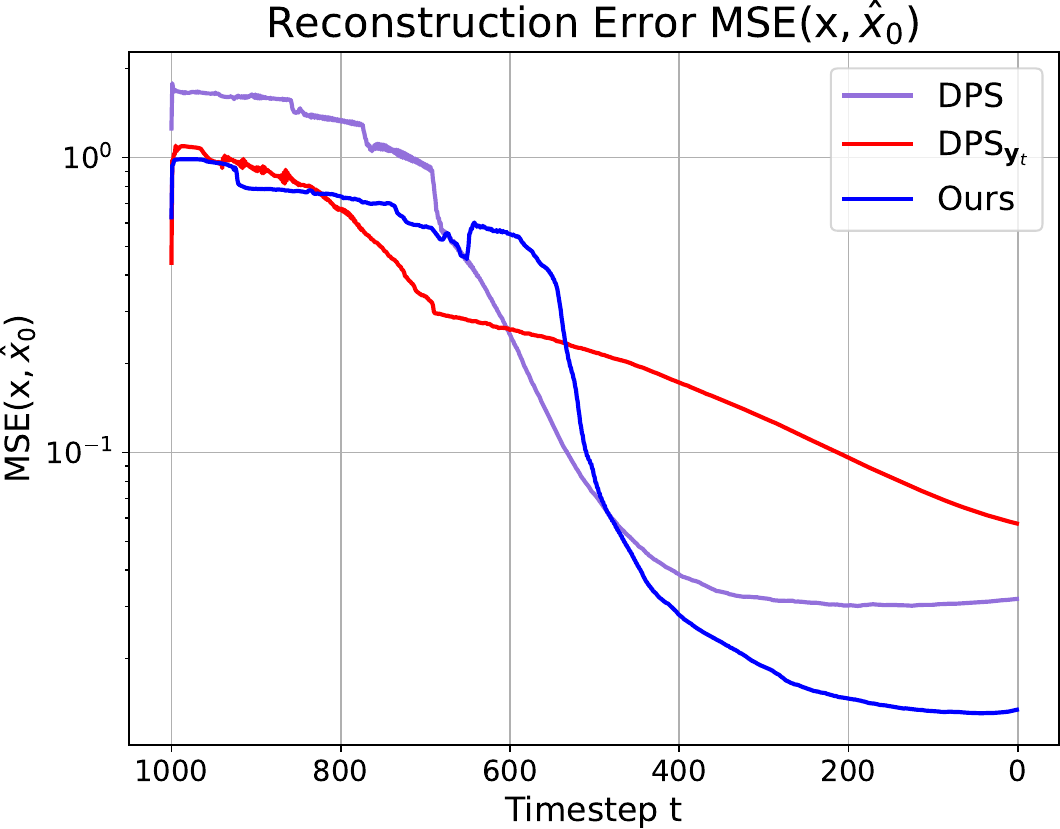}
        \caption{}
        \label{fig:reconstruction}
    \end{subfigure}
    
    \vspace{-0.2cm}
    \caption{(a) mean magnitude ratio between high- and low-frequency area at timestep $t$; (b) $\epsilon$-prediction error ratio: $\frac{\text{DPS}_{\boldsymbol{y}_t}}{\text{DPS}}$ and $\frac{\text{DPS-CM}}{\text{DPS}}$ at timestep $t$; (c) reconstruction error between target $\boldsymbol{x}$ and intermediate $\hat{\boldsymbol{x}}_0(\boldsymbol{x}_t)$ at timestep $t$. These results are achieved by performing Gaussian Deblurring on 10 FFHQ images.}
    \vspace{-0.2cm}
    \label{fig:motivation}
\end{figure*}
Our claim is that solving inverse problems in the posterior sampling manner follows similar frequency dynamics with the regular denoising process in that it recovers the low-frequency (structural, domain-independent) components at first and the high-frequency (details, domain-dependent) components later. The posterior sampling from $ p_t\left(\boldsymbol{x}_t| \boldsymbol{y}\right)$ in DPS is inevitable adding high-frequency signals from the gradient of the approximate log-likelihood $\nabla_{\boldsymbol{x}_t} \log p_t\left(\boldsymbol{y} | \hat{\boldsymbol{x}}_0\right)$ during the early stages due to the frequency and noise discrepancy between the sharp measurement $\boldsymbol{y}$ and intermediate noisy reconstruction $\hat{\boldsymbol{x}}_0$. It will create a distorted frequency dynamic which leads to a larger posterior sampling error. 

To verify our assumption, we conduct frequency dynamic analysis of different posterior sampling methods for solving inverse problems in Fig.\ref{fig:freq}. Afterward, the posterior sampling error is illustrated via the visualization of the $\epsilon$-prediction error and intermediate reconstruction error dynamics in Fig.\ref{fig:epsilon} and Fig.\ref{fig:reconstruction} respectively. We conduct the diffusion posterior sampling from $p_t\left(\boldsymbol{x}_t| \boldsymbol{y}_t\right)$  with noisy measurement $\text{DPS}_{\boldsymbol{y}_t}$ in comparison with DPS to illustrate the advantage of the likelihood approximation with less high-frequency elements.

\textbf{Diffusion posterior sampling with the noisy measurement $\text{DPS}_{\boldsymbol{y}_t}$.} Given the noisy measurement $\boldsymbol{y}_t$ constructed by the diffusion forward process starting from the clean measurement $\boldsymbol{y}$ as in Eq. \ref{eq:formxt},  
we replace $\boldsymbol{y}$ with $\boldsymbol{y}_t$ to form $\text{DPS}_{\boldsymbol{y}_t}$ with the posterior sampling from $p_t\left(\boldsymbol{x}_t| \boldsymbol{y}_t\right)$. The corresponding likelihood approximation $p_t\left(\boldsymbol{y}_t| \hat{\boldsymbol{x}}_0\right)\simeq p_t\left(\boldsymbol{y}_t| \boldsymbol{x}_t\right)$ involves $\boldsymbol{y}_t$ and $\boldsymbol{x}_t$ with the similar frequency pattern and noise level.

\textbf{Frequency dynamic of diffusion posterior sampling.} We verify our assumption on frequency dynamic of diffusion posterior sampling by transforming the gradient of log posterior (i.e., $\nabla_{\boldsymbol{x}_t} \log p\left(\boldsymbol{x}_t | \boldsymbol{y}\right)$ in the case of DPS) into spectral space as the gradient of log posterior serves as the main update for each posterior sampling step in Eq.\ref{eq:con}. We split the low-frequency and high-frequency areas in the spectral space of the gradient with the low-high cut-off frequency of 32Hz and visualize the mean magnitude ratio between them of each $t$ shown in Fig.\ref{fig:freq}.

\textbf{$\epsilon$-prediction error and intermediate reconstruction error.} Since there exists no ground truth for the intermediate generation of diffusion posterior sampling for inverse problems, we measure the posterior sampling error implicitly by computing $\epsilon$-prediction error and intermediate reconstruction error at timestep $t$. Given intermediate $\boldsymbol{x}_t$ and the posterior mean $\hat{\boldsymbol{x}}_0$ by applying Eq.\ref{eq:x0hat}, the $\epsilon$-prediction error is formulated as $\left\|\epsilon_\theta\left(\boldsymbol{x}'_t,t\right)-\epsilon\right\|^2_2$, where $\boldsymbol{x}'_t=\sqrt{\bar{\alpha}_t} \hat{\boldsymbol{x}}_0+\sqrt{1-\bar{\alpha}_t} \epsilon, \epsilon \sim \mathcal{N}(\mathbf{0}, \boldsymbol{I})$ and $\epsilon_\theta\left(\boldsymbol{x}'_t,t\right)=-\sqrt{1-\bar{\alpha}_t}\boldsymbol{s}_\theta\left(\boldsymbol{x}'_t, t\right)$ by applying Tweedie’s formula. $\epsilon$-prediction error at timestep $t$ can reflect how accurately the posterior sampling recovers at $t$. We compare different posterior sampling with DPS by computing their ratio of $\epsilon$-prediction error at each timestep shown in Fig.\ref{fig:epsilon}, where a ratio below 1 indicates a smaller $\epsilon$-prediction error than DPS, and above 1 means a larger one. Besides, the reconstruction error formed as $\text{MSE}(\boldsymbol{x},\hat{\boldsymbol{x}}_0)$ is shown in Fig.\ref{fig:reconstruction} to evaluate how well the intermediate reconstruction $\hat{\boldsymbol{x}}_0(\boldsymbol{x}_t)$ converges to the target $\boldsymbol{x}$ utilizing the mean squared error.

\textbf{Observations.} From Fig.\ref{fig:freq}, the frequency dynamic of both DPS's $\nabla_{\boldsymbol{x}_t} \log p\left(\boldsymbol{x}_t | \boldsymbol{y}\right)$ (purple) and $\text{DPS}_{\boldsymbol{y}_t}$'s $\nabla_{\boldsymbol{x}_t} \log p\left(\boldsymbol{x}_t | \boldsymbol{y}_t\right)$ (red) focuses on low-frequency components and gradually recovers more on high-frequency components, but $\text{DPS}_{\boldsymbol{y}_t}$ maintains larger attention on the low-frequency recovery throughout all timesteps. This difference arises from the discrepancy between $\boldsymbol{y}$ and $\boldsymbol{y}_t$ and leads to their contrast in $\epsilon$-prediction and reconstruction error. From timestep 1000 to 600 (around) as the early stages of diffusion posterior sampling, $\text{DPS}_{\boldsymbol{y}_t}$ prohibits a smaller $\epsilon$-prediction error than DPS (the red curve of Fig.\ref{fig:epsilon}). Consistently accompanying it, the reconstruction error of $\text{DPS}_{\boldsymbol{y}_t}$ is smaller and converges faster before $t=600$ by comparing DPS (purple) and $\text{DPS}_{\boldsymbol{y}_t}$ (red) in Fig.\ref{fig:reconstruction}. 

Such observations verify our claim that the gradient of log-likelihood $\nabla_{\boldsymbol{x}_t} \log p\left(\boldsymbol{y} | \hat{\boldsymbol{x}}_0\right)$ involved in DPS introduces more high-frequency signals during early stages which poses negative effects on early-time low-frequency recovery reflecting from the larger posterior sampling error shown in DPS. The success of $\text{DPS}_{\boldsymbol{y}_t}$ in early stages arises from the similar noise and frequency level between $\boldsymbol{y}_t$ and $\hat{\boldsymbol{x}}_0$ which brings less biased guidance for low-frequency generations before $t=600$.
\begin{figure}[t]
    \centering
    \includegraphics[width=1.0\linewidth]{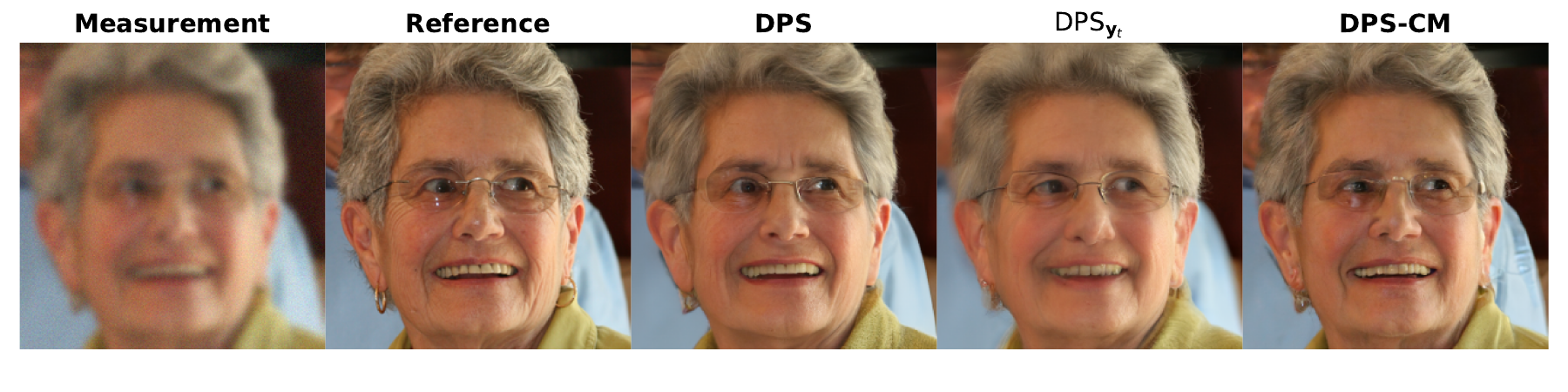}
    \vspace{-1cm}
    \caption{\textbf{Qualitative example of Gaussian Deblurring on FFHQ dataset including $\text{DPS}_{\boldsymbol{y}_t}$.} 
    All measurements are corrupted by additional Gaussian noise with a variance of $\sigma_{\boldsymbol{y}}=0.05$. 
    }
    \vspace{-1cm}
    \label{fig:yt}
\end{figure}
However, from $t=600$ to $0$, when the diffusion model gradually transfers its focus from low-frequency to high-frequency restoration shown in Fig.\ref{fig:freq}, the random noise $\boldsymbol{u}$ in $\boldsymbol{y}_t$ brings guidance mismatching the real details, which leads $\text{DPS}_{\boldsymbol{y}_t}$ with a larger posterior sampling error than DPS shown in Fig.\ref{fig:epsilon} and \ref{fig:reconstruction}. Further, in Figure \ref{fig:yt}, $\text{DPS}_{\boldsymbol{y}_t}$'s generation lacks details and looks ambiguous compared with DPS. 

The proposed DPS-CM preserves the advantage of $\text{DPS}_{\boldsymbol{y}_t}$ at the early stages of sampling but still keeps accurate high-frequency generations during the later period. DPS-CM leverages the crafted measurement $\mathbf{y}_t$ instead of random sampled $\boldsymbol{y}_t$ to perform preciser posterior sampling from $\nabla_{\boldsymbol{x}_t} \log p\left(\boldsymbol{x}_t | \mathbf{y}_t\right)$, while $\mathbf{y}_t$ is sampled from another diffusion denoising process. 
We will formulate DPS-CM in detail and discuss its advantages in \ref{sec:3_2}.
\subsection{Enhancing Posterior Sampling by Integrating Crafted Noisy Measurements}\label{sec:3_2}
$\text{DPS}_{\boldsymbol{y}_t}$ introduced in \ref{sec:3_1} fits the diffusion model $\epsilon_{\theta}(\cdot,t)$ adaptively and produces smaller $\epsilon$-prediction errors, but fails to recover high-frequency components (details, domain-dependent signals)  during the late denoising stage. Besides the random noise obstructing the generation of real details, this failure also origins from the intractable likelihood term $p\left(\boldsymbol{y}_t | \boldsymbol{x}_t\right)$ in Eq.\ref{eq:dpsyt}:
\begin{equation}
\begin{aligned}
\nabla_{\boldsymbol{x}_t} \log p\left(\boldsymbol{x}_t | \boldsymbol{y}_t\right)&=\nabla_{\boldsymbol{x}_t} \log p\left(\boldsymbol{x}_t\right)+\nabla_{\boldsymbol{x}_t} \log \textcolor{blue}{p\left(\boldsymbol{y}_t | \boldsymbol{x}_t\right)}.
\label{eq:dpsyt}
\end{aligned}
\end{equation}
Apparently, there exists no explicit dependency between the noisy measurement $\boldsymbol{y}_t$ and $\boldsymbol{x}_t$ nor $\hat{\boldsymbol{x}}_0$ applying Eq.\ref{eq:x0hat} given the measurement model. Thus, the $\nabla_{\boldsymbol{x}_t} \log p\left(\boldsymbol{y}_t | \hat{\boldsymbol{x}}_0\right)$ in $\text{DPS}_{\boldsymbol{y}_t}$ is an inadequate estimate for $\nabla_{\boldsymbol{x}_t} \log p\left(\boldsymbol{x}_t | \boldsymbol{y}_t\right)$,
 but provides benefits during early restoration as shown in \ref{sec:3_1}. To take advantage of the tractable measurement model $p\left(\boldsymbol{y} | \boldsymbol{x}_0\right)$ with the involvement of $\boldsymbol{y}_t$, the gradient of log posterior in $\text{DPS}_{\boldsymbol{y}_t}$ can be factorized as: 
\[
\begin{aligned}
\nabla_{\boldsymbol{x}_t} \log p\left(\boldsymbol{x}_t | \boldsymbol{y}_t\right)
&=\nabla_{\boldsymbol{x}_t} 
\log \int p\left(\boldsymbol{x}_t | \boldsymbol{y}_0, \boldsymbol{y}_t\right) 
       p\left(\boldsymbol{y}_0 | \boldsymbol{y}_t\right) d \boldsymbol{y}_0 \\
&=\nabla_{\boldsymbol{x}_t} 
\log \int p\left(\boldsymbol{x}_t | \boldsymbol{y}_0\right) 
       p\left(\boldsymbol{y}_0 | \boldsymbol{y}_t\right) 
\, d \boldsymbol{y}_0 \\
&=\nabla_{\boldsymbol{x}_t} 
\log \mathbb{E}_{\boldsymbol{y}_0 \sim p\left(\boldsymbol{y}_0 | \boldsymbol{y}_t\right)}
     \bigl[p\left(\boldsymbol{x}_t | \boldsymbol{y}_0\right)\bigr]\\
&=\nabla_{\boldsymbol{x}_t} \log p\left(\boldsymbol{x}_t \right) \\
&\quad + \nabla_{\boldsymbol{x}_t} \log
  \mathbb{E}_{\boldsymbol{y}_0 \sim p\left(\boldsymbol{y}_0 | \boldsymbol{y}_t\right)}
  \bigl[p\left(\boldsymbol{y}_0 | \boldsymbol{x}_t\right)\bigr].
\end{aligned}
\]
With a similar factorization on posterior $p\left(\boldsymbol{y}_0 | \boldsymbol{x}_t\right)$, we have: 
\begin{equation}
\begin{aligned}
\nabla_{\boldsymbol{x}_t} &\log\mathbb{E}_{\boldsymbol{y}_0 \sim p\left(\boldsymbol{y}_0 | \boldsymbol{y}_t\right)}\left[p\left(\boldsymbol{y}_0 | \boldsymbol{x}_t\right)\right] \\
&= \nabla_{\boldsymbol{x}_t} \log\mathbb{E}_{\boldsymbol{y}_0 \sim p\left(\boldsymbol{y}_0 | \boldsymbol{y}_t\right)}[\mathbb{E}_{\boldsymbol{x}_0 \sim p\left(\boldsymbol{x}_0 | \boldsymbol{x}_t\right)}[p\left(\boldsymbol{y}_0 | \boldsymbol{x}_0\right)]]\\
&\simeq\nabla_{\boldsymbol{x}_t} \log\mathbb{E}_{\boldsymbol{y}_0 \sim p\left(\boldsymbol{y}_0 | \boldsymbol{y}_t\right)}\underbrace{[p\left(\boldsymbol{y}_0 | \hat{\boldsymbol{x}}_0\right)]}_{\text{DPS approximation}}.
\end{aligned}
\label{eq:fact}
\end{equation}
where $p\left(\boldsymbol{y}_0 | \boldsymbol{x}_t\right)$ is approximated as $p\left(\boldsymbol{y}_0 | \hat{\boldsymbol{x}}_0\right)$ in DPS, given the posterior mean $\hat{\boldsymbol{x}}_0$ in Eq.\ref{eq:x0hat} and the input measurement $\boldsymbol{y}$ as $\boldsymbol{y}_0$. The failure of $\text{DPS}_{\boldsymbol{y}_t}$ comes from the highly biased approximation of $\mathbb{E}\left[\boldsymbol{y}_0 | \boldsymbol{y}_t\right]$ as $\boldsymbol{y}_t$ to form $\nabla_{\boldsymbol{x}_t} \log p_t\left(\boldsymbol{y}_t | \hat{\boldsymbol{x}}_0\right)$ in sampling.
According to the observations in \ref{sec:3_1}, both cases are practically flawed: DPS increases posterior errors during early stages and $\text{DPS}_{\boldsymbol{y}_t}$ produces biased high-frequency recovery in the later period. 

To retain their benefits while alleviating the drawbacks, we propose to craft noisy measurements $\boldsymbol{y}_t$ from a diffusion denoising sampling so that we can utilize $\mathbb{E}\left[\boldsymbol{y}_0 | \boldsymbol{y}_t\right]$ in the case of VP-SDE to move the outer expectation of $p\left(\boldsymbol{y}_0 | \hat{\boldsymbol{x}}_0\right)$ over the posterior distribution $\mathbb{E}\left[\boldsymbol{y}_0 | \boldsymbol{y}_t\right]$ into the inner expectation of $\boldsymbol{y}_0$. To distinguish it from $\text{DPS}_{\boldsymbol{y}_t}$, we use $\mathbf{y}_t$ to denote crafted noisy measurements rather than $\boldsymbol{y}_t$:
\begin{equation}
\begin{aligned}
\nabla_{\boldsymbol{x}_t} &\log\mathbb{E}_{\mathbf{y}_0 \sim p\left(\mathbf{y}_0 | \mathbf{y}_t\right)}\left[p\left(\mathbf{y}_0 | \boldsymbol{x}_t\right)\right] \\
&\simeq\nabla_{\boldsymbol{x}_t} \log\mathbb{E}_{\mathbf{y}_0 \sim p\left(\mathbf{y}_0 | \mathbf{y}_t\right)}[p\left(\mathbf{y}_0 | \hat{\boldsymbol{x}}_0\right)]\\
&\simeq\nabla_{\boldsymbol{x}_t}\log \underbrace{p\left(\hat{\mathbf{y}}_0 | \hat{\boldsymbol{x}}_0\right)}_{\text{Our approximation}}.
\end{aligned}
\label{eq:approx1}
\end{equation}
where $\hat{\mathbf{y}}_0:=\mathbb{E}\left[\mathbf{y}_0 | \mathbf{y}_t\right]=\mathbb{E}_{\mathbf{y}_0 \sim p\left(\mathbf{y}_0 | \mathbf{y}_t\right)}[\mathbf{y}_0]$ in Eq.\ref{eq:x0hat}.

\textbf{Step 1: craft the noisy measurement $\mathbf{y}_t$.} 
There are ways to craft target noisy measurement $\mathbf{y}_t$ for our approximation in Eq.\ref{eq:approx1} for the case of VP-SDE: via diffusion denoising sampling or inversion of diffusion models~\citep{mokady2023null}. One issue for crafting $\mathbf{y}_t$ is there is no off-the-shelf model for the measurement($\boldsymbol{y}$) modality. While $\boldsymbol{x}$ and $\boldsymbol{y}$ lie on close manifolds in typical inverse problems, we continue to utilize the trained score estimator $\boldsymbol{s}_\theta(\cdot, t)$ on $\boldsymbol{x}$ for the craft measurement $\mathbf{y}_t$. Due to the accumulation of errors throughout diffusion inversion timesteps~\citep{huberman2024edit}, the crafted measurement trajectory $\{\mathbf{y}_t\}_{t=T}^0$ is not constructed via a diffusion inversion process but a diffusion denoising sampling on $\mathbf{y}_t$ conditioned on the given measurement $\boldsymbol{y}$ with the gradient of the posterior $\nabla_{\mathbf{y}_t} \log p\left(\mathbf{y}_t | \boldsymbol{y}\right)$. Similar to the approximation in Eq.\ref{eq:post} in the case of Gaussian noise, the posterior $\nabla_{\mathbf{y}_t} \log p\left(\mathbf{y}_t | \boldsymbol{y}\right)$ is formed as: 
\begin{equation}
    \nabla_{\mathbf{y}_t} \log p_t\left(\mathbf{y}_t | \boldsymbol{y}\right) \simeq \boldsymbol{s}_{\theta}\left(\mathbf{y}_t, t\right)-\rho \nabla_{\mathbf{y}_t}\left\|\hat{\mathbf{y}}_0-\boldsymbol{y}\right\|_2^2.
\end{equation}
Specifically, $\mathbf{y}_t$ is updated iteratively as \textbf{lines 3-6} as Step 1 in Algorithm \ref{alg:dps-cm}. Throughout the trajectory, $\mathbf{y}_t$ is gradually remedied to the vanilla measurement $\boldsymbol{y}$. 

\textbf{Step 2: integrate crafted measurements $\mathbf{y}_t$.} 
We incorporate the crafted $\mathbf{y}_t$ in Step 1 into our approximated log-likelihood gradient of Eq.\ref{eq:approx1} to form DPS-CM. Unlike the biased $\boldsymbol{y}_t$ as $\mathbb{E}\left[\boldsymbol{y}_0 | \boldsymbol{y}_t\right]$ in $\text{DPS}_{\boldsymbol{y}_t}$, the $\hat{\mathbf{y}}_0 =\mathbb{E}\left[\mathbf{y}_0 | \mathbf{y}_t\right]$ with gradually denoised crafted measurement $\mathbf{y}_t$ will not cause information loss but inject signals adaptively from low-frequency to high-frequency matching the restoration denoising trajectory of $\boldsymbol{x}$ with the close model distribution. In each iteration, after the update of the crafted measurement $\mathbf{y}_t$, leveraging the approximation in Eq.\ref{eq:approx1}, it is clear to construct $\nabla_{\boldsymbol{x}_t} \log p\left(\boldsymbol{x}_t | \mathbf{y}_t\right)$ in the case of Gaussian noise as follows: 
\begin{equation}
\nabla_{\boldsymbol{x}_t} \log p_t\left(\boldsymbol{x}_t | \mathbf{y}_t\right) \simeq \boldsymbol{s}_{\theta}\left(\boldsymbol{x}_t, t\right)-\zeta_{t} \nabla_{\boldsymbol{x}_t}\left\|\hat{\mathbf{y}}_0-\mathcal{A}\left(\hat{\boldsymbol{x}}_0\right)\right\|_2^2.
\label{eq:cm}
\end{equation}
We perform the diffusion posterior sampling incorporating the estimate in Eq.\ref{eq:cm} with crafted measurements on Gaussian deblurring and visualize the frequency pattern and sampling error in Fig.\ref{fig:motivation}. 
As shown in Fig.\ref{fig:freq}, DPS-CM (blue) concentrates more on the low-frequency recovery in the early stages compared with DPS and $\text{DPS}_{\boldsymbol{y}_t}$, and adjusts itself to high-frequency components better than $\text{DPS}_{\boldsymbol{y}_t}$ during the late period. Besides, in Fig.\ref{fig:epsilon} and \ref{fig:reconstruction}, DPS-CM (blue) has the overall smallest $\epsilon$-prediction error and achieves the best final generation closest to ground truth with the smallest reconstruction error.
Empirically, combining DPS with DPS-CM leads to even better results and benefits the sampling to jump out of the plateau in the early stages of sampling. Thus, we integrate the approximated gradient of log-likelihood $\nabla_{\boldsymbol{x}_t} \log p\left(\boldsymbol{y} | \hat{\boldsymbol{x}}_0\right)$ of DPS with our proposed $\nabla_{\boldsymbol{x}_t} \log p\left(\hat{\mathbf{y}}_0 | \hat{\boldsymbol{x}}_0\right)$ to facilitate inverse problem solving. The final diffusion reverse sampling of $\boldsymbol{x}$'s restoration in DPS-CM is shown as \textbf{lines 7-10} as Step 2 in Algorithm \ref{alg:dps-cm}. In Appendix~\ref{app:possion}, DPS-CM for measurements with Poisson noise is shown in Algorithm \ref{alg:dps-cm-poisson}.
\begin{algorithm}[H]
\caption{DPS-CM}
\label{alg:dps-cm}
\begin{algorithmic}[1]
\Require Forward operator $\mathcal{A}(\cdot)$, $T$, \textcolor{blue}{Measurement $\boldsymbol{y}$}, Step size $\left\{\omega_t\right\}_{t=1}^T$, $\left\{\zeta_t\right\}_{t=1}^T$ for $\mathbf{y}$ and $\boldsymbol{x}$, $\mu\in [0,1]$ for integration control, Pretrained score estimator $\boldsymbol{s}_{\theta}(\cdot, t)$
\State $\mathbf{y}_T \sim \mathcal{N}(\mathbf{0}, \boldsymbol{I})$, $\boldsymbol{x}_T \sim \mathcal{N}(\mathbf{0}, \boldsymbol{I})$ 
\For{$t = T-1 $ to $1$}
    \State $\boldsymbol{s}_\mathbf{y} \gets \boldsymbol{s}_{\theta}(\mathbf{y}_{t}, t)$ \Comment{Start of $\mathbf{y}$ trajectory (\textbf{Step 1})}
    \State $\hat{\mathbf{y}}_0 \leftarrow \frac{1}{\sqrt{\bar{\alpha}_t}}\left(\mathbf{y}_t+\left(1-\bar{\alpha}_t\right) \boldsymbol{s}_\mathbf{y}\right)$  
    
    \Comment{Estimate of $\mathbb{E}\left[\mathbf{y}_0 | \mathbf{y}_t\right]$}
    \State $\mathbf{y}'_{t-1} \gets \frac{1}{\sqrt{1-\beta_t}}\left(\mathbf{y}_t+\beta_t \boldsymbol{s}_\mathbf{y}\right) + \sigma_t z_{\mathbf{y}}$ 
    \State $\mathbf{y}_{t-1} \gets \mathbf{y}'_{t-1}-  \omega_{t}\nabla_{\mathbf{y}_t} \|
    \hat{\mathbf{y}}_0-\textcolor{blue}{\boldsymbol{y}}\|_2$  
    
    \Comment{Posterior Sampling for $\mathbf{y}$}
    \State $\boldsymbol{s}_{\boldsymbol{x}} \gets \boldsymbol{s}_{\theta}(\boldsymbol{x}_{t}, t)$ \Comment{Start of $\boldsymbol{x}$ trajectory (\textbf{Step 2})}
    \State $\hat{\boldsymbol{x}}_0 \leftarrow \frac{1}{\sqrt{\bar{\alpha}_t}}\left(\boldsymbol{x}_t+\left(1-\bar{\alpha}_t\right) \boldsymbol{s}_{\boldsymbol{x}}\right)$  
    
    \Comment{Estimate of $\mathbb{E}\left[\boldsymbol{x}_0 | \boldsymbol{x}_t\right]$}
    \State $\boldsymbol{x}'_{t-1} \gets \frac{1}{\sqrt{1-\beta_t}}\left(\boldsymbol{x}_t+\beta_t \boldsymbol{s}_{\boldsymbol{x}}\right) + \sigma_t z_{\boldsymbol{x}}$ 
    \State $\boldsymbol{x}_{t-1} \gets \boldsymbol{x}'_{t-1}- \zeta_{t} \nabla_{\boldsymbol{x}_t} (\mu\cdot\|\hat{\mathbf{y}}_0-\mathcal{A}\left(\hat{\boldsymbol{x}}_0\right)\|_2+(1-\mu)\cdot\|\textcolor{blue}{\boldsymbol{y}}-\mathcal{A}\left(\hat{\boldsymbol{x}}_0\right)\|_2)$  \Comment{Integrate Crafted $\hat{\mathbf{y}}_0$}
\EndFor
\State \Return $\mathbf{x}_0$
\end{algorithmic}
\end{algorithm}

%% file: main/experiments.tex
\section{Experiments}
\subsection{Experimental Settings}
\textbf{Evaluation datasets and baselines.} We evaluate our proposed posterior sampling method DPS-CM on the validation set of FFHQ 256$\times$256 and ImageNet 256$\times$256 datasets by applying DDPM sampling with 1000 timesteps. For the diffusion prior, the pre-trained diffusion models for FFHQ and ImageNet are taken from DPS \citep{chung2022diffusion} and ablated diffusion model (ADM) \citep{dhariwal2021diffusion} respectively. Detailed hyperparameter $\{\zeta_t,\omega_t,\mu\}$ settings of DPS-CM are shown in Appendix \ref{app:hyper}. We compare our method with several recent state-of-the-art approaches including DPS \citep{chung2022diffusion}, Denoising Diffusion Restoration Models (DDRM) \citep{kawar2022denoising}, DiffPIR \citep{zhu2023denoising}, Optimal Posterior Covariance (OPC) \citep{peng2024improving}, and also latent diffusion-based methods: Posterior Sampling with Latent Diffusion (PSLD) \citep{rout2024solving} and ReSample \citep{song2023solving}. We also compare DPS-CM with FPS-SMC \citep{dou2023diffusion} and LGD-MC \citep{song2023loss} for ablation study in Section.\ref{sec:ablation} as they also improve posterior sampling with augmented $\boldsymbol{y}_t$ or $\boldsymbol{x}$. Although FPS-SMC relies on the separable Gaussian kernel for deblurring and LGD-MC is applied for more general tasks, they are still worthy of comparison in investigating the effect from different posterior estimate designs. We utilize the same pre-trained models used in DPS-CM for baselines DPS, DDRM, OPC, DiffPIR, FPS-SMC, and LGD-MC. For PSLD, Stable Diffusion v-1.5 \citep{rombach2022high} is applied. For ReSample, we use the VQ-4 autoencoder and the FFHQ-LDM with CelebA-LDM from latent diffusion \citep{rombach2022high}. For OPC, we report the performance of the convert posterior covariance version with Type I guidance, as this version has the best and most stable performance among all variants. We assume that all the measurements are injected in Gaussian noise with standard deviation $\sigma=0.05$. For measurements with Poisson noise, the noise level is $\lambda=1.0$. We run our experiments on a single GPU Nvidia A6000. We list the baseline settings in detail in the Appendix \ref{app:baseline}.  
\begin{figure}[tb]  
    \centering
    \begin{subfigure}[b]{0.497\textwidth}
        \includegraphics[width=\textwidth]{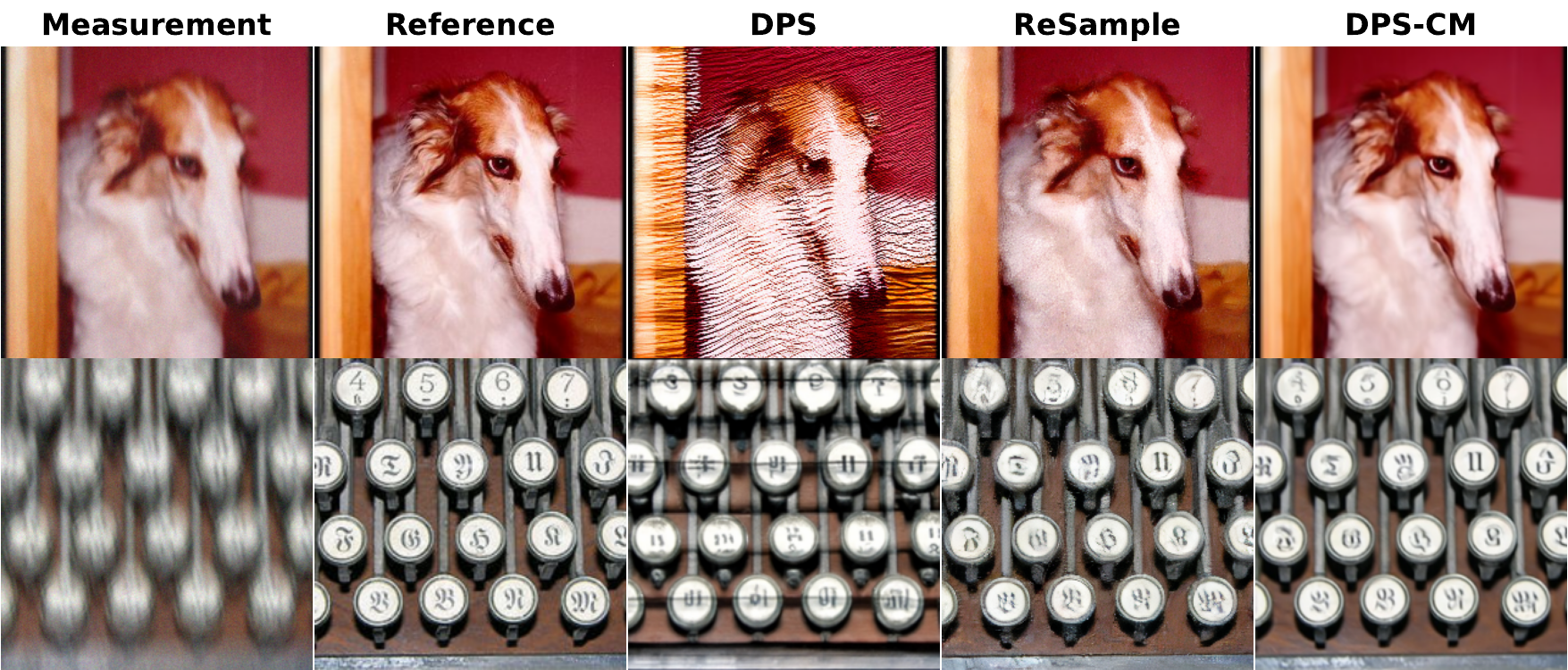}
        \caption{Motion Deblurring}
        \label{fig:md}
    \end{subfigure}
    \begin{subfigure}[b]{0.497\textwidth}  
        \includegraphics[width=\textwidth]{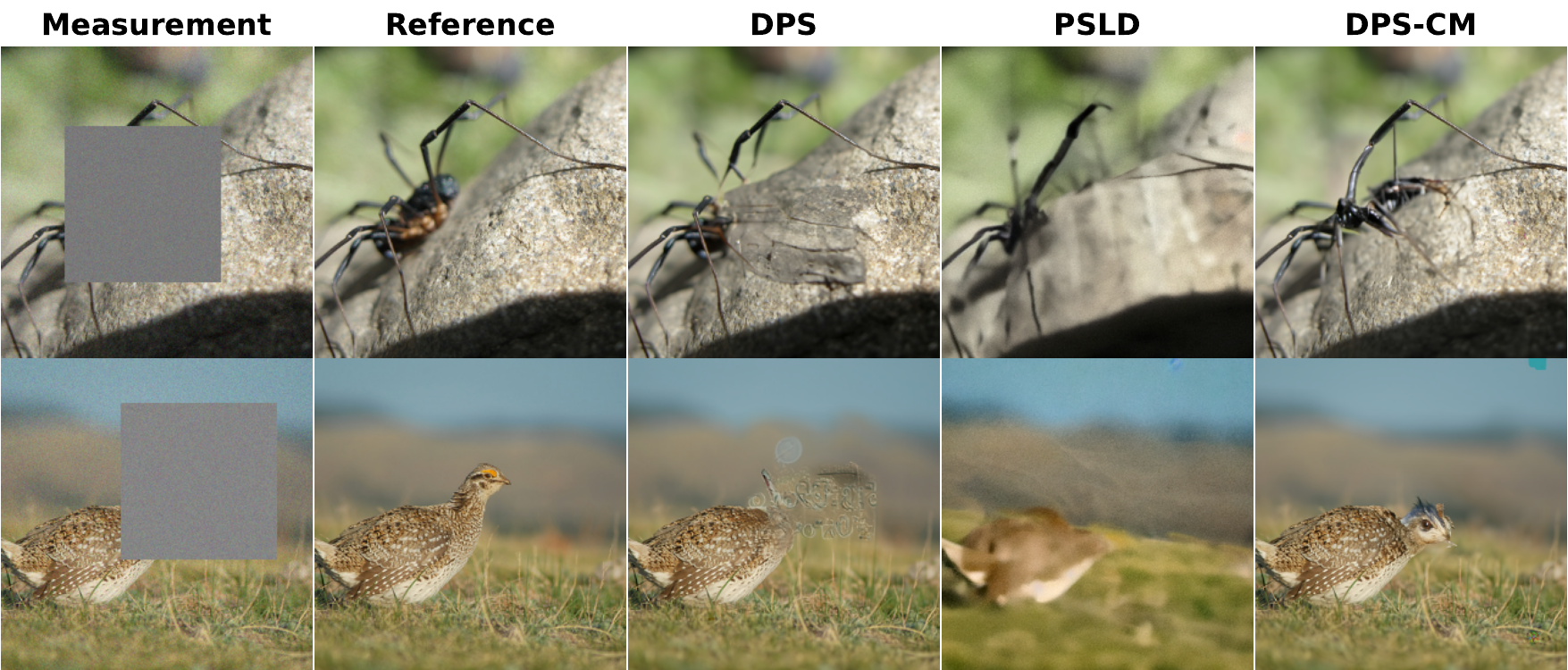}
        \caption{Box Inpainting}
        \label{fig:bi}
    \end{subfigure}
    \vspace{-0.8cm}
    \caption{\textbf{Qualitative results of motion deblurring and inpainting (box) on ImageNet dataset.} All measurements are corrupted by additional Gaussian noise with a variance of $\sigma_{\boldsymbol{y}}=0.05$.
    }
    \vspace{-0.5cm}
    \label{fig:deblur}
\end{figure}

\textbf{Experimental setting for forward operators.} The forward measurement operators are mostly following \cite{chung2022diffusion}: (1) for random inpainting, 70 percent of image pixels are masked in all channels; (2) for box inpainting, a box region with size 128$\times$128 in an image is masked; (3) for Gaussian blur, we use the same operator implemented in DPS \citep{chung2022diffusion} with kernel size 61$\times$61 and standard deviation value 3.0; (4) We take the implementation from \cite{LeviBorodenko} to form motion blur operator with kernel size 61$\times$61 and intensity value 0.5; (5) For 4$\times$ super-resolution, we consider bicubic downsampling method; (6) For nonlinear deblurring, we utilize the approximated
forward model in \cite{tran2021explore} to simulate the real-world blur.  
\begin{table*}[!t]
\renewcommand{\baselinestretch}{1.0}
\renewcommand{\arraystretch}{1.0}
\setlength{\tabcolsep}{2pt}
\centering
\small
\resizebox{\textwidth}{!}{
\begin{tabular}{@{}cccccccccccccc@{}}
\toprule
\multirow{2}{*}{\textbf{Dataset}} & \multirow{2}{*}{\textbf{Method}}      & \multicolumn{4}{c}{\textbf{Gaussian Deblur}} & \multicolumn{4}{c}{\textbf{Motion Deblur}} & \multicolumn{4}{c}{\textbf{super-resolution~($4\times$)}}\\
\cmidrule(lr){3-6}  \cmidrule(lr){7-10}  \cmidrule(lr){11-14} 
&    & PSNR~$\uparrow$ & SSIM~$\uparrow$ & LPIPS~$\downarrow$ & FID~$\downarrow$& PSNR~$\uparrow$ & SSIM~$\uparrow$ & LPIPS~$\downarrow$ & FID~$\downarrow$& PSNR~$\uparrow$ & SSIM~$\uparrow$ & LPIPS~$\downarrow$ & FID~$\downarrow$ \\
\midrule
\multirow{6}{*}{FFHQ}     & DPS-CM~(\textit{Ours}) &  \underline{27.45}&\textbf{0.7819}& \textbf{0.2059}&\textbf{56.04} &\textbf{30.31}& \textbf{0.8585} & \textbf{0.1609}&\textbf{42.58}& \underline{27.81} &  \textbf{0.7942}& \textbf{0.2008} & \textbf{51.72} \\
                          & DDRM  & 25.20 &0.7289 &0.2976&105.06 & - &- &-&-&26.89 &0.7821 &0.2667&95.23\\
                          & DPS & 26.35 & 0.7533&\underline{0.2178}&65.20& 27.84 & \underline{0.8023}&\underline{0.2129}&70.10& 27.55 & \underline{0.7886}&\underline{0.2077}&\underline{57.20}\\
                          & OPC &  27.06& 0.7651&0.2196&\underline{61.76}&  26.20&0.7371&0.2535&\underline{66.63} &  26.90& 0.7637&0.2400&76.33\\
                          & DiffPIR &  24.13& 0.6649&0.2919&76.96&  26.35&0.7394&0.2487&64.25 & 25.74& 0.7143&0.2711&67.19\\
\cmidrule(lr){2-14}
                          & PSLD &  \textbf{28.45}&  \underline{0.7689}& 0.3019& 99.90&  26.19&  0.6779& 0.3667& 137.88&  \textbf{27.91}&  0.7783& 0.2783& 87.82 \\
                          & ReSample &26.10&  0.6441& 0.3361& 100.21&\underline{28.26}&  0.7221& 0.2937& 89.80&23.23&  0.4517& 0.5061& 151.20\\
\midrule 
\multirow{6}{*}{ImageNet} & DPS-CM~(\textit{Ours})   & \underline{22.73} & \textbf{0.6147}    & \textbf{0.3397}    & \textbf{128.92}& \textbf{24.36} & \textbf{0.6814} & \textbf{0.3163} & \textbf{97.54}    &\underline{23.78}&\textbf{0.6450}&\textbf{0.3242}&\textbf{97.38}\\
                          & DDRM  & 21.48 &0.5527 &0.4948&242.38 & - &- &-&-&22.72 &0.6225 &0.4252&196.32\\
                          & DPS          & 16.36 &0.3449&0.5041&208.49 & 17.41 & 0.3970&0.4920&204.27& 21.17 & 0.5404&0.3668&106.71\\
                          & OPC   &  18.93&  0.4254&0.4579&132.38 &  18.43&0.3782&0.4989&174.44 &  19.44& 0.4113&0.4836&169.71\\
                          & DiffPIR  &  20.62&  0.4753&0.4445&154.12 &23.38&0.6285&0.3711&120.58 &  22.97& 0.5985&0.3840&119.66\\
\cmidrule(lr){2-14}
                          & PSLD &  \textbf{23.45}&  \underline{0.6089}& \underline{0.3419}& \underline{131.90}&  23.18&  0.5688& 0.4206& 167.57& \textbf{23.79}&  \underline{0.6371}& \underline{0.3346}& \underline{120.68} \\
                          & ReSample&22.79&  0.5147& 0.4355& 172.17 &\underline{23.95}&  \underline{0.5723}& \underline{0.3929}& \underline{125.33}&21.04&  0.3973& 0.5001& 203.56\\
\bottomrule
\end{tabular}}
\vspace{-8pt}
\caption{\textbf{Quantitative results for delur and super-resolution on FFHQ and ImageNet dataset.} We use \textbf{bold} and \underline{underline} to indicate the best and second-best results, respectively.}
\vspace{-6pt}
\label{tab:gdmdsr}
\end{table*}
\textbf{Evaluation metrics.} We report the average performance of 100 validation samples of 4 metrics: peak signal-to-noise ratio (PSNR), structural similarity index (SSIM), Learned Perceptual Patch Similarity (LPIPS) \citep{dosovitskiy2016generating}, and Fréchet Inception Distance (FID) \citep{heusel2017gans} for comprehensive evaluations. 
\subsection{Quantitative Results}
The quantitative results of DPS-CM and baselines on these noisy linear and nonlinear inverse problems are shown in Table~\ref{tab:gdmdsr},~\ref{tab:inpaint} , and Table~\ref{tab:nd}. Besides, the performance of DPS-CM for measurements with Poisson noise is shown in Table~\ref{tab:poi} of Appendix \ref{app:possion}. We can observe that DPS-CM can achieve the overall best performance over four metrics when solving deblurring and super-resolution problems and significantly outperform baselines on random/box inpainting problems. While the performance and stability of DPS will significantly drop when performing Gaussian/Motion deblurring on the ImageNet dataset, DPS-CM with the crafted measurements can instead form more precise and robust posterior sampling reflected from the consistent performance over these two datasets. Optimal Posterior Covariance (OPC), an improved posterior estimate by constructing the optimal posterior covariance instead of improving over the expectation, is no better than DPS-CM but still achieves comparable performance with our method. For latent diffusion-based baselines PSLD and Resample, It is interesting that they can handle the more challenging inverse problems on the ImageNet better than the FFHQ dataset, which exhibits the advantage of latent diffusion capturing on more complex visual distributions and semantic patterns. Remarkably, our method exceeds baselines with different diffusion priors when solving motion deblurring problems, shown in examples of Figure \ref{fig:md}. Besides, DPS-CM outperforms DPS in solving ill-posed problems, such as nonlinear blurring and inverse problems with Poisson noise as shown in Table~\ref{tab:nd} and Appendix \ref{app:possion}. In Fig.\ref{fig:deblur}, \ref{fig:gd}, \ref{fig:ri} , and \ref{fig:sr}, compared with baselines on different noisy linear inverse problems, DPS-CM shows high-quality reconstructions with mild distortion and precise detail recovery. For example, DPS-CM has better letter recovery on the keyboards in Fig.\ref{fig:md} and signature recovery (under the car) in Fig.\ref{fig:ri}. In super-resolution and box inpainting, DPS-CM displays fewer missing details and provides reconstructions suited to the natural background and masked objects. For nonlinear deblurring in Fig.\ref{fig:nb}, DPS reconstructs measurements to the target to some extent but with extra noise and distortions injected, and DPS-CM can capture the realistic details close to the target in this nonlinear task. Additional visual examples of DPS-CM and baselines on different inverse problems with Gaussian and Poisson noise are included in Appendix \ref{app:visual}.
\subsection{Ablation Studies}\label{sec:ablation}
\begin{wrapfigure}[8]{r}[0pt]{0.23\textwidth} 
\vspace{-35pt}
\caption{Ablation study on $\mu$.}
\includegraphics[width=\linewidth]{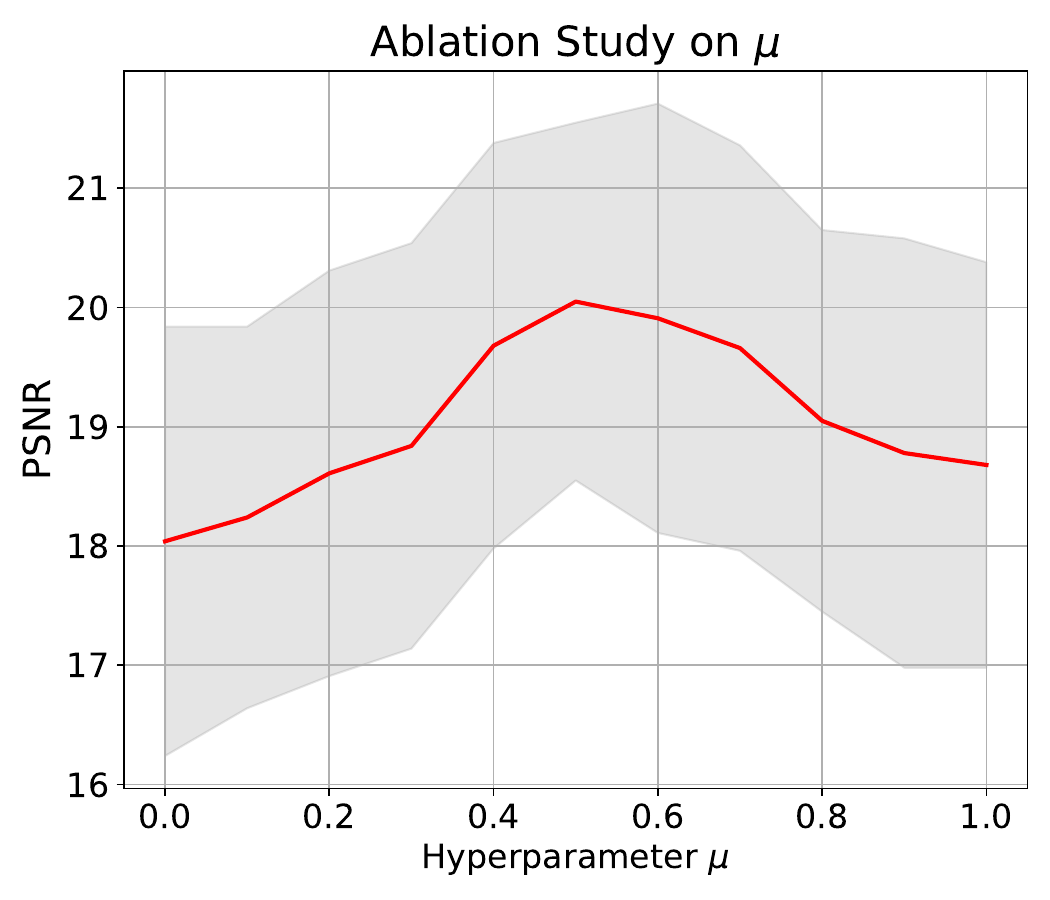} 
\label{fig:abla}
\end{wrapfigure}
\textbf{Influence of hyperparameter $\mu$.} Here, we want to investigate the influence of hyperparameter $\mu\in [0,1]$, which controls the proportion of proposed $\nabla_{\boldsymbol{x}_t} \log p_t\left(\boldsymbol{y}_t \mid \mathbf{x}_t\right)$ in the sampling of DPS-CM. When $\mu=0.0$, it falls into DPS. Specifically, we can observe the performance (PSNR) of DPS-CM on box inpainting on ImageNet with different $\mu$ in Figure~\ref{fig:abla}. DPS-CM achieves the best results when $\mu=0.5$ which indicates the benefits of the interplay between $\nabla_{\boldsymbol{x}_t} \log p_t\left(\mathbf{y}_t \mid \mathbf{x}_t\right)$ and $\nabla_{\boldsymbol{x}_t} \log p_t\left(\boldsymbol{y} \mid \mathbf{x}_t\right)$. Besides, pure DPS-CM with $\mu=1.0$ leads to a better log posterior gradient estimate and outperforms DPS-CM with $\mu=0.0$.
\begin{table*}[!h]
\renewcommand{\baselinestretch}{1.0}
\renewcommand{\arraystretch}{1.0}
\setlength{\tabcolsep}{2pt}
\centering
\small
\begin{tabular}{@{}cccccccccc@{}}
\toprule
\multirow{2}{*}{\textbf{Dataset}} & \multirow{2}{*}{\textbf{Method}}      & \multicolumn{4}{c}{\textbf{Inpainting~(Random)}} & \multicolumn{4}{c}{\textbf{Inpainting~(Box)}} \\
\cmidrule(lr){3-6}  \cmidrule(lr){7-10}  
&    & PSNR~$\uparrow$ & SSIM~$\uparrow$ & LPIPS~$\downarrow$ & FID~$\downarrow$& PSNR~$\uparrow$ & SSIM~$\uparrow$ & LPIPS~$\downarrow$ & FID~$\downarrow$ \\
\midrule
\multirow{6}{*}{FFHQ}     & DPS-CM~(\textit{Ours})   &  \textbf{31.20}  &  \textbf{0.8903} & \textbf{0.1233}  &  \textbf{28.05}&\textbf{23.87}  &\textbf{0.8470}  & \textbf{0.1331} &\textbf{35.52}    \\
                          & DDRM  & 25.49 & 0.7727 &0.2702 &101.26&  20.40&  0.8005& 0.2265& 65.12\\
                          & DPS & 29.74 & 0.8136&\underline{0.1397}&\underline{43.12} & 21.54 & 0.8127&\underline{0.1514}&50.69\\
                          & OPC &  \underline{29.84}& \underline{0.8709}&0.1615&50.49&  \underline{22.69}& \underline{0.8222}&0.1569&\underline{48.01}\\
                          & DiffPIR &29.40& 0.8425&0.2275&83.09&21.95&0.7253&0.2484&63.40\\
\cmidrule(lr){2-10}
                          & PSLD &  28.52&  0.7756& 0.3165& 101.04&  21.07&  0.6894& 0.3761& 128.24\\
                          & ReSample&29.65&  0.8367& 0.2024& 61.44&19.69&  0.7837& 0.2452& 136.84 \\
\midrule 
\multirow{6}{*}{ImageNet} & DPS-CM~(\textit{Ours})   & \textbf{26.71} & \textbf{0.8121}    & \textbf{0.1684}    & \textbf{37.92} & \textbf{20.05} & \textbf{0.7880} & \textbf{0.2270} & \textbf{111.01}    \\
                          & DDRM  & 21.89& 0.6204 &0.4162 &230.95 &  \underline{18.32}&  0.7373& 0.3003& 142.95\\
                          & DPS & \underline{26.61} & \underline{0.8060}&\underline{0.1824}&\underline{39.92}  & 18.04 & \underline{0.7626}&\underline{0.2201}&\underline{119.87}\\
                          & OPC&  23.46&  0.6517&0.3063&88.47&  16.46& 0.6412&0.2910&168.26\\
                          & DiffPIR &25.79& 0.7470&0.2873&106.79&18.08&0.6030&0.3326&160.95\\
\cmidrule(lr){2-10}
                          & PSLD &  25.46&  0.6841& 0.3509& 104.99& 18.17&  0.5469& 0.4701& 220.96\\
                          & ReSample &25.01&  0.7201& 0.2825& 98.45&17.50&  0.6862& 0.3346& 225.58 \\
\bottomrule
\end{tabular}
\vspace{-6pt}
\caption{\textbf{Quantitative results for inpainting on FFHQ and ImageNet dataset.} We use \textbf{bold} and \underline{underline} to indicate the best and second best results, respectively.}
\vspace{-10pt}
\label{tab:inpaint}
\end{table*}
\begin{figure}[htbp] 
\centering 
\begin{minipage}{\linewidth} 
\centering 
\small
\resizebox{0.9\textwidth}{!}{
\setlength{\tabcolsep}{2pt}
\renewcommand{\arraystretch}{1.0}
\begin{tabular}{@{}ccccccc@{}}
\toprule
 & \multirow{2}{*}{\textbf{Method}} & \multicolumn{4}{c}{\textbf{Nonlinear Deblur}} \\
\cmidrule(lr){3-6}
 & & PSNR~$\uparrow$ & SSIM~$\uparrow$ & LPIPS~$\downarrow$ & FID~$\downarrow$ \\
\midrule
 & DPS-CM~(\textit{Ours}) & \textbf{22.66} & \textbf{0.6465} & \textbf{0.3626} & \textbf{122.64} \\
 & $\text{DPS}$ & 21.21& 0.6191& 0.3846& 136.86\\
\bottomrule
\end{tabular}}
\vspace{-0.15cm}
\captionof{table}{Quantitative results for nonlinear deblurring on FFHQ.}
\label{tab:nd}
\end{minipage}%
\vspace{-0.5cm}
\end{figure}

{\bf Compare crafted measurements with other augmented posterior estimate methods.}
Here, we further validate the effect of our crafted measurements $\left\{\mathbf{y}_t\right\}_{t=T}^0$ compared with $\text{DPS}_{\boldsymbol{y}_t}$, where $\boldsymbol{y}_t$ in the posterior is an i.i.d sample from the forward process $p\left(\boldsymbol{y}_t \mid \boldsymbol{y}_0\right)$ same as Eq.\ref{eq:formxt}, LGD-MC \citep{song2023loss}, which is a loss guided diffusion with $\boldsymbol{x}^{(i)}$ i.i.d.~sampled from $ p\left(\mathbf{x}_0 \mid \mathbf{x}_t\right)$ for the posterior estimate, and FPS-SMC which incorporates the $\mathbf{y}_{t-1}$ sampled from the non-parametric forward/reverse process given $\mathbf{y}$. Specifically, in the inverse problem setting, the gradient of the log-likelihood LGD-MC is approximated as:
\[
\mathrm{MC}_n\left(\mathbf{x}_t, \mathbf{y}\right)=\rho \nabla_{\mathbf{x}_t} \log (\frac{1}{n} \sum_{i=1}^n\left\|\boldsymbol{y}-\mathcal{A}\left(\hat{\boldsymbol{x}}_0^{(i)}\right)\right\|_2^2),
\]
where $\boldsymbol{x}_0^{(i)}\sim p\left(\boldsymbol{x}_0 \mid \boldsymbol{x}_t\right)$. $\text{DPS}_{\boldsymbol{y}_t}$ with multiple Monte-Carlo samples can be formed as:
\[
\nabla_{\boldsymbol{x}_t} \log p\left(\boldsymbol{y}_t \mid \boldsymbol{x}_t\right)=\rho \nabla_{\mathbf{x}_t} \log (\frac{1}{n} \sum_{i=1}^n\left\|\boldsymbol{y}_t^{(i)}-\mathcal{A}\left(\hat{\boldsymbol{x}}_0\right)\right\|_2^2),
\]
where $\boldsymbol{y}_t^{(i)}\sim p\left(\boldsymbol{y}_t \mid \boldsymbol{y}_0\right)$. And the estimate in FPS-SMC is $p_{\boldsymbol{\theta}}\left(\mathbf{y}_{t-1} \mid \mathbf{x}_{t-1}\right)$ given sampled $\mathbf{y}_{t-1}$. DPS-CM improves the posterior estimate differently via crafted measurements. We conduct experiments on 4$\times$ super-resolution to demonstrate the superiority of our crafted measurements $\mathbf{y}_t$ over these augmented posterior estimate methods. The Monte Carlo sample number is set as 10 for all of them ($M=10$ for FPS-SMC). Shown in Table \ref{tab:abla1}, DPS-CM shows the best results on perception-oriented metrics (LPIPS and FID) and comparable performance on the standard metrics (PSNR and SSIM) with FPS-SMC. DPS-CM is 50$\%$ more efficient than FPS-SMC as shown in the running time report in Appendix \ref{app:runtime}, which validates the design in this work can keep the details and promote posterior approximation at the same time. More visual comparisons with FPS-SMC are shown in Fig.\ref{fig:appendix_sr_fps} and \ref{fig:sr}.
\begin{figure}[t] 
\centering 
\begin{minipage}{\linewidth} 
\centering 
\small
\resizebox{0.9\textwidth}{!}{
\setlength{\tabcolsep}{2pt}
\renewcommand{\arraystretch}{1.0}
\begin{tabular}{@{}ccccccc@{}}
\toprule
 & \multirow{2}{*}{\textbf{Method}} & \multicolumn{4}{c}{\textbf{super-resolution~($4\times$)}} \\
\cmidrule(lr){3-6}
 & & PSNR~$\uparrow$ & SSIM~$\uparrow$ & LPIPS~$\downarrow$ & FID~$\downarrow$ \\
\midrule
 & DPS-CM~(\textit{Ours}) & \underline{27.81}& \underline{0.7942} & \textbf{0.2008} & \textbf{51.72} \\
 & $\text{DPS}_{\boldsymbol{y}_t}$ & 22.40& 0.6514& 0.3273& 96.34\\
 & LGD-MC &26.35&0.7674 & 0.2359&61.98 \\
 & FPS-SMC &\textbf{28.12}&\textbf{0.8053} & \underline{0.2140}&\underline{60.22} \\
\bottomrule
\end{tabular}
}
\vspace{-0.15cm}
\captionof{table}{Ablation study results on FFHQ comparing crafted measurements in DPS-CM with other augmented posterior estimate methods.}
\vspace{-0.5cm}
\label{tab:abla1}
\end{minipage}%
\end{figure}

%% file: main/conlusion.tex
\section{Conclusion}
This work introduces Diffusion Posterior Sampling with Crafted Measurements to solve general and noisy inverse problems. DPS-CM leverages an additional diffusion reverse process to form a crafted measurement trajectory $\{\mathbf{y}_t\}_{t=T}^0$ with extensive noise levels to perform a frequency-adaptive and less biased diffusion posterior sampling with the log posterior gradient $\nabla_{\boldsymbol{x}_t} \log p\left(\boldsymbol{x}_t \mid \mathbf{y}_t\right)$ incorporating $\mathbf{y}_t$. Via experimental results, we show that our DPS-CM can generate improved high-quality restorations compared with recent state-of-the-art and also latent diffusion approaches. For potential future works, following the idea of DPS-CM, improved design on forming more beneficial crafted measurements, e.g., additional inverse problem-related guidance for $\mathbf{y}_t$ diffusion sampling besides the reconstruction loss guidance,  is a crucial direction to explore. We discuss the limitations of DPS-CM and solutions in Appendix~\ref{app:limit}.

%% file: main/appendix.tex
\newpage
\appendix
\onecolumn
\section{Algorithm and results: DPS-CM for measurements with Poisson noise}\label{app:possion}
DPS-CM is applied in the case of the measurements with Poisson noise in Algorithm \ref{alg:dps-cm-poisson}. Its performance on the deblurring and super-resolution is shown in Table.\ref{tab:poi}.
\begin{center}
\begin{minipage}{1.0\linewidth}
\begin{algorithm}[H]
\caption{DPS-CM (Poisson)}
\label{alg:dps-cm-poisson}
\begin{algorithmic}[1]
\Require Forward operator $\mathcal{A}(\cdot)$, $T$, \textcolor{blue}{Measurement $\boldsymbol{y}$}, Step size for $\mathbf{y}$: $\left\{\omega_t\right\}_{t=1}^T$, Step size for $\boldsymbol{x}$: $\left\{\zeta_t\right\}_{t=1}^T$, Hyperparameter $\mu\in [0,1]$ for integration control, Score function $\boldsymbol{s}_{\theta}(\cdot, t)$
\State $\mathbf{y}_T \sim \mathcal{N}(\mathbf{0}, \boldsymbol{I})$, $\boldsymbol{x}_T \sim \mathcal{N}(\mathbf{0}, \boldsymbol{I})$ 
\For{$t = T-1 $ to $1$}
    \Statex
    \State $\boldsymbol{s}_\mathbf{y} \gets \boldsymbol{s}_{\theta}(\mathbf{y}_{t}, t)$ \Comment{Start of $\mathbf{y}$ trajectory}
    \State$\hat{\mathbf{y}}_0 \leftarrow \frac{1}{\sqrt{\bar{\alpha}_t}}\left(\mathbf{y}_t+\left(1-\bar{\alpha}_t\right) \boldsymbol{s}_\mathbf{y}\right)$ \Comment{Mean Estimate of $\mathbb{E}\left[\mathbf{y}_0 \mid \mathbf{y}_t\right]$}
    \State $\mathbf{y}'_{t-1} \gets \frac{1}{\sqrt{1-\beta_t}}\left(\mathbf{y}_t+\beta_t \boldsymbol{s}_\mathbf{y}\right) + \sigma_t z_{\mathbf{y}}$ \Comment{Unconditional DDPM Sampling of $\mathbf{y}$}
    \State $\mathbf{y}_{t-1} \gets \mathbf{y}'_{t-1}-\omega_t \nabla_{\mathbf{y}_t}\left\|\hat{\mathbf{y}}_0-\textcolor{blue}{\boldsymbol{y}}\right\|_{\boldsymbol{\Lambda}}$ \Comment{Reconstruction Loss Guidance}
    \Statex
    \State $\boldsymbol{s}_{\boldsymbol{x}} \gets \boldsymbol{s}_{\theta}(\boldsymbol{x}_{t}, t)$ \Comment{Start of $\boldsymbol{x}$ trajectory}
    \State$\hat{\boldsymbol{x}}_0 \leftarrow \frac{1}{\sqrt{\bar{\alpha}_t}}\left(\boldsymbol{x}_t+\left(1-\bar{\alpha}_t\right) \boldsymbol{s}_{\boldsymbol{x}}\right)$ \Comment{Mean Estimate of $\mathbb{E}\left[\boldsymbol{x}_0 \mid \boldsymbol{x}_t\right]$}
    \State $\boldsymbol{x}'_{t-1} \gets \frac{1}{\sqrt{1-\beta_t}}\left(\boldsymbol{x}_t+\beta_t \boldsymbol{s}_{\boldsymbol{x}}\right) + \sigma_t z_{\boldsymbol{x}}$ \Comment{Unconditional DDPM Sampling of $\boldsymbol{x}$}
    \State $\boldsymbol{x}_{t-1} \gets \boldsymbol{x}'_{t-1}-\zeta_t \nabla_{\boldsymbol{x}_t}\left(\mu\left\|\hat{\mathbf{y}}_0-\mathcal{A}\left(\hat{\boldsymbol{x}}_0\right)\right\|_{\boldsymbol{\Lambda}}+(1-\mu)\left\|\textcolor{blue}{\boldsymbol{y}}-\mathcal{A}\left(\hat{\boldsymbol{x}}_0\right)\right\|_{\boldsymbol{\Lambda}}\right.$ \Comment{Integrating Crafted $\hat{\mathbf{y}}_0$}
\EndFor
\State \Return $\mathbf{x}_0$
\end{algorithmic}
\end{algorithm}
\end{minipage}
\end{center}
\begin{table*}[h]
\renewcommand{\baselinestretch}{1.0}
\renewcommand{\arraystretch}{1.0}
\setlength{\tabcolsep}{2pt}
\centering
\small
\resizebox{\textwidth}{!}{
\begin{tabular}{@{}cccccccccccccc@{}}
\toprule
\multirow{2}{*}{\textbf{Dataset}} & \multirow{2}{*}{\textbf{Method}}      & \multicolumn{4}{c}{\textbf{Gaussian Deblur}} & \multicolumn{4}{c}{\textbf{Motion Deblur}} & \multicolumn{4}{c}{\textbf{super-resolution~($4\times$)}}\\
\cmidrule(lr){3-6}  \cmidrule(lr){7-10}  \cmidrule(lr){11-14} 
&    & PSNR~$\uparrow$ & SSIM~$\uparrow$ & LPIPS~$\downarrow$ & FID~$\downarrow$& PSNR~$\uparrow$ & SSIM~$\uparrow$ & LPIPS~$\downarrow$ & FID~$\downarrow$& PSNR~$\uparrow$ & SSIM~$\uparrow$ & LPIPS~$\downarrow$ & FID~$\downarrow$ \\
\midrule
\multirow{2}{*}{FFHQ}     & DPS-CM~(\textit{Ours}) &  \textbf{26.23}&\textbf{0.7368}& \textbf{0.2348}&\textbf{66.61} &\textbf{27.53}& \textbf{0.7687} & \textbf{0.2208}&\textbf{53.19}& \textbf{26.40} &  \textbf{0.7417}& \textbf{0.2552} & \textbf{72.97} \\
                         & DPS & 25.22 & 0.7184&0.2391&71.82& 27.01 &0.7608&0.2213&58.78& 25.17 &0.6708&0.3391&106.53\\
\bottomrule
\end{tabular}}
\caption{Quantitative results for deblurring and super-resolution on FFHQ with \textbf{Poisson noise.} We use \textbf{bold} for the best.}
\vspace{-6pt}
\label{tab:poi}
\end{table*}

\section{Experimental Details}
\subsection{Implementation settings of DPS-CM} \label{app:hyper}
Hyperparameters $\{\zeta_t,\omega_t,\mu\}$ for DPS-CM are fixed during the 1000 DDPM denoising steps. Detailed settings in the case of Gaussian and Poisson noise are shown in Table ~\ref{tab:hyperparameter_dps_cm_gau} and ~\ref{tab:hyperparameter_dps_cm_poi} respectively.
\begin{table}[!h]
    \centering
    \resizebox{1.0\textwidth}{!}{%
    \begin{tabular}{ccccccc}
    \toprule 
         & Gaussian Deblur  & Motion Deblur&super-resolution&Inpainting(Random)&Inpainting(Box)&Nonlinear Deblur\\\midrule
         FFHQ& $\{1.8, 13.0 ,0.5\}$&$\{1.7 ,11.0, 0.5\}$&$\{2.2, 8.0, 0.5\}$&$\{3.1, 19.0 ,0.285\}$&$\{1.7 ,13.0, 0.5\}$& $\{1.6, 10.0, 0.5\}$\\
         ImageNet& $\{1.8, 13.0 ,0.7\}$&$\{1.4 ,11.0 ,0.5\}$&$\{1.7, 7.0, 0.5\}$&$\{2.4 ,19.0 ,0.285\}$&$\{1.8 ,13.0 ,0.5\}$&-\\\bottomrule
    \end{tabular}}
    \caption{Hyperparameter settings of DPS-CM for measurements with Gaussian noise}
    \label{tab:hyperparameter_dps_cm_gau}
\end{table}

\begin{table}[!h]
    \centering
    \resizebox{0.5\textwidth}{!}{%
    \begin{tabular}{ccccccc}
    \toprule 
         & Gaussian Deblur  & Motion Deblur&super-resolution\\\midrule
         FFHQ& $\{0.5, 6.0 ,0.5\}$&$\{0.4 ,6.0, 0.5\}$&$\{0.4, 3.0, 0.5\}$
         \\\bottomrule
    \end{tabular}}
    \caption{Hyperparameter settings of DPS-CM for measurements with Poisson noise}
    \label{tab:hyperparameter_dps_cm_poi}
\end{table}
\subsection{Implementation settings of baselines} \label{app:baseline}
\textbf{DPS}: we follow all the default $\rho$ settings in Appendix D.1 in their paper with 1000 DDPM steps.

\textbf{DDRM}: the number of steps we used is 20. DDRM with 100 steps sometimes has better results on PSNR and SSIM but worse on LPIPS and FID. The overall performance between different step schedules is very similar. For hyperparameters $\{\eta,\eta_b\}$, we follow the default setting $\{0.85,1\}$.

\textbf{OPC}: we compare their different variants and report the Convert version with their defined Type I guidance which has the overall best performance with 50 EDM steps.

\textbf{DiffPIR}: we follow the hyperparameter settings in Appendix B.1 with 100 timesteps. 

\textbf{FPS-SMC}: particle size $M$ is 10 and the $c$ value is followed by the setting in their Table 8.

\textbf{LGD-MC}: LGD-MC is performed with $n=10$. The loss function follows the settings in Appendix B.2 with $r_t=0.05$, the loss coefficient $\lambda=10^{-3}$ and 1000 timesteps.

\textbf{PSLD}: we follow the default hyperparameter setting $\{\eta=1, \gamma=0.1\}$ and sample for 1000 DDIM steps.

\textbf{ReSample}: we implement it with the default settings of their paper and codebase: $\{\tau=10^{-4},\gamma=40\}$, conducting latent optimization with a maximum of 500 iterations and sampling for 500 DDIM steps.

\section{Running Time of DPS-CM and baselines}\label{app:runtime}
We show the running time/NFEs of DPS-CM and baselines in Table~\ref{tab:sample_speed}.
\begin{table}[!h]
    \centering
    \resizebox{0.45\textwidth}{!}{%
    \begin{tabular}{ccc}
    \toprule 
         Methods& Running Time(in seconds)  & NFEs\\\midrule
         DPS-CM& $123.22$&$2000$ \\
         DDRM& $0.54$&$20$ \\
         DPS& $64.70$&$1000$\\
         OPC& $9.69$&$50$ \\
         DiffDIR& $3.92$&$100$ \\
         LGD-MC& 66.45&$1000$ \\
         FPS-SMC& $182.67$&$1000$\\
         PSLD& $517.47$&$1000$ \\
         ReSample& $639.51$&$500$\\\bottomrule
    \end{tabular}}
    \vspace{10pt}
    \caption{Implementation details of DPS-CM}
    \label{tab:sample_speed}
\end{table}
\section{Limitations}\label{app:limit} 

As the high-quality generations of DPS-CM benefit from the interplay between two posterior sampling processes for the crafted measurement's generation and $\boldsymbol{x}$'s restoration, the running time of DPS-CM is nearly twice of DPS's sampling as shown in Tabel~\ref{tab:sample_speed}. However, DPS-CM shows better or comparable performance compared with latent diffusion-based methods such as PSLD, Resample, and methods with Monte Carlo such as FPS-SMC and LGD-MC with less running time. Besides, we try to improve the efficiency of DPS-CM by setting $\mu=0$ after $t$ during the late stages of sampling as the posterior sampling with crafted measurements shows similar $\epsilon$-prediction errors with DPS during the late stages in Fig.~\ref{fig:epsilon}. We set the $t$ as 400 and conduct an experiment for random inpainting on the FFHQ dataset. From Table~\ref{tab:improved}, the accelerated variant achieves comparable performance and has over \textbf{20$\%$ efficiency improvement}. Besides, the core idea of DPS-CM is based on insightful observations in posterior sampling methods. We can potentially combine crafted measurements with a more efficient method, such as DiffPIR, to achieve high-quality reconstructions and maintain efficiency. Another concern for DPS-CM is how it deals with the inverse problems with the targets and measurements from very different modalities, such as phase retrieval. In this case, utilizing the same pre-trained diffusion model to construct the crafted measurements for DPS-CM sampling can not generate stable restorations. One solution is to train a smaller diffusion model on the measurement modality for crated measurements. It should be noted that most of the current zero-shot methods are designed for typical inverse problems in which the target and measurement lie on the close manifold and perform unstably on tasks similar to phase retriever. Thus, fitting a small model to facilitate DPS-CM for such tasks is acceptable and reasonable. 
\begin{table*}[t]
\renewcommand{\baselinestretch}{1.0}
\renewcommand{\arraystretch}{1.0}
\setlength{\tabcolsep}{2pt}
\centering
\small
\resizebox{0.5\textwidth}{!}{
\begin{tabular}{@{}cccccc@{}}
\toprule
\multirow{2}{*}{\textbf{Running Time}} & \multirow{2}{*}{\textbf{Method}}      & \multicolumn{4}{c}{\textbf{super-resolution~($4\times$)}}\\
\cmidrule(lr){3-6} 
&    & PSNR~$\uparrow$ & SSIM~$\uparrow$ & LPIPS~$\downarrow$ & FID~$\downarrow$ \\
\midrule
123.22     & DPS-CM &  \textbf{31.20}&\underline{0.8903}& \textbf{0.1233}&\textbf{28.05} \\
                        98.14 & DPS-CM (accelerated) & \underline{30.98} & \textbf{0.8906}&\underline{0.1349}& \underline{36.64}\\
                         64.70& DPS & 29.74 & 0.8136&0.1397&43.12\\
\bottomrule
\end{tabular}}
\caption{Quantitative results for random inpainting on FFHQ with Gaussian noise of \textbf{accelerated} DPS-CM}
\vspace{-6pt}
\label{tab:improved}
\end{table*}
\section{Additional Visual Results} \label{app:visual}
\begin{figure}[H] 
    \centering
    \includegraphics[width=1\linewidth]{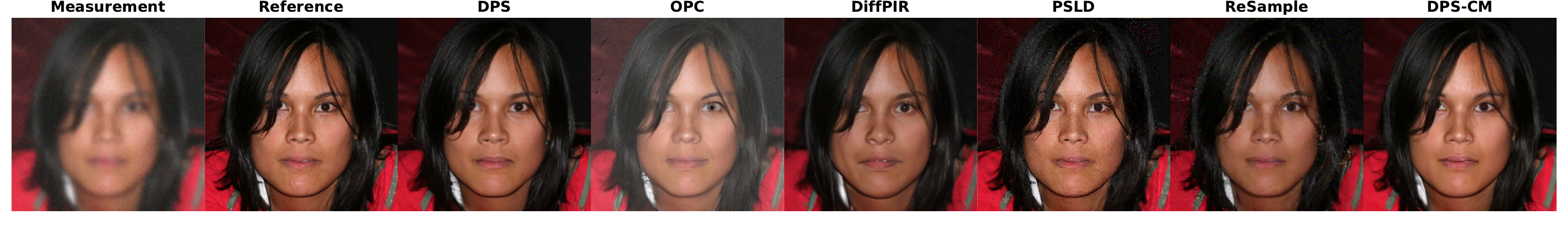}
    \caption{Additional Qualitative Results on Gaussian Deblur with Gaussian Noise.}
    \label{fig:appendix_gd}
    \vspace{-15pt} 
\end{figure}

\begin{figure}[H] 
    \centering
    \includegraphics[width=0.7\linewidth]{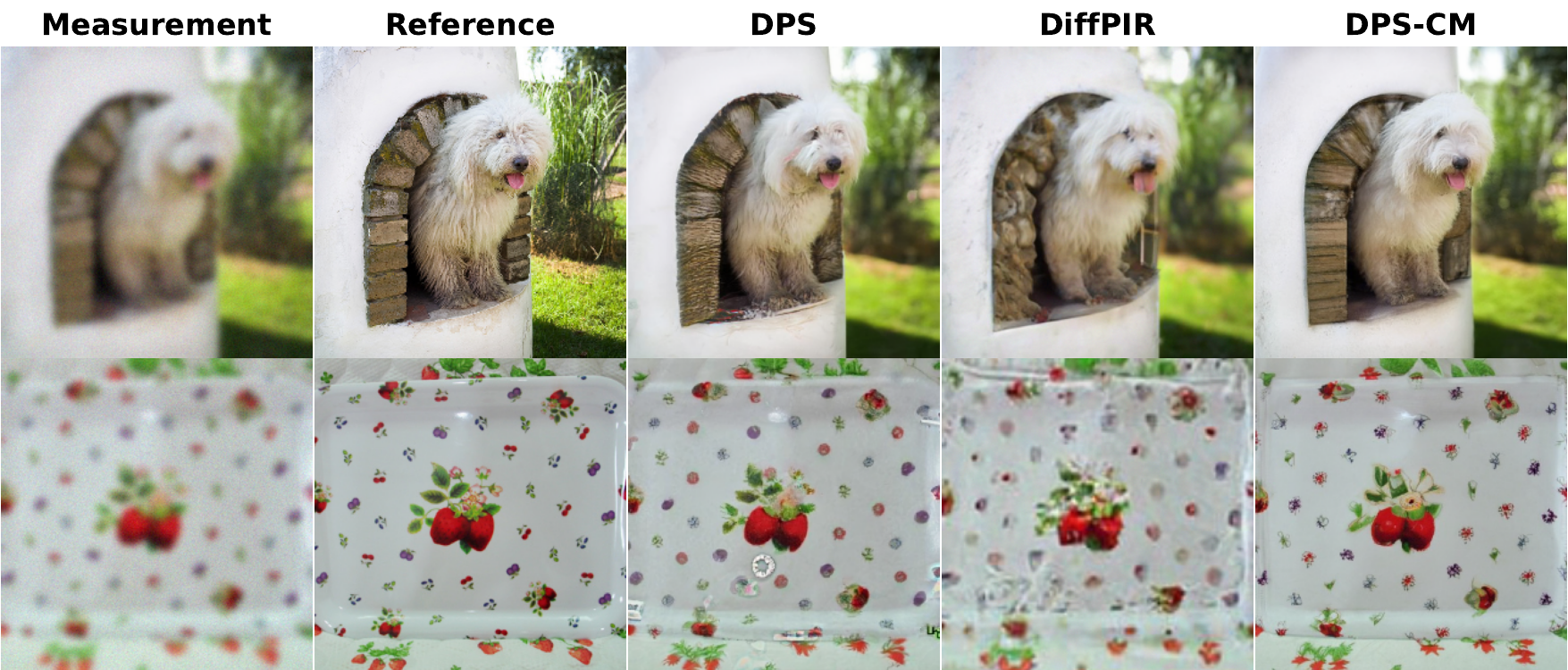}
    \caption{Additional Qualitative Results on Gaussian Deblur with Gaussian Noise.}
    \label{fig:gd}
    \vspace{-15pt} 
\end{figure}

\begin{figure}[H] 
    \centering
    \includegraphics[width=1\linewidth]{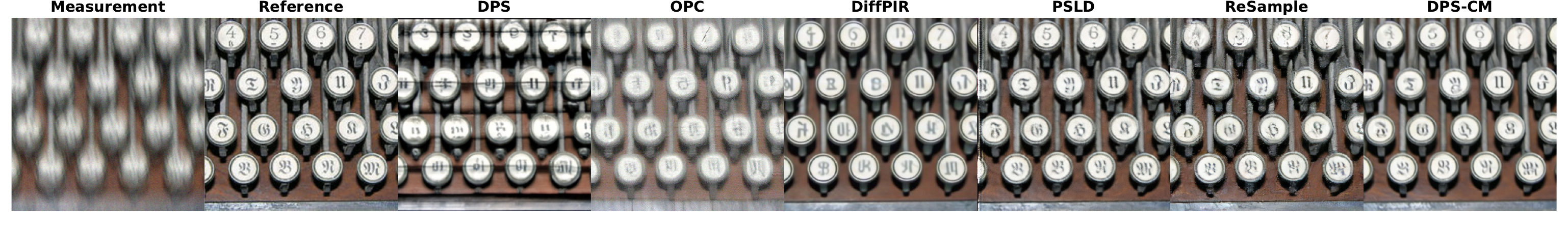}
    \caption{Additional Qualitative Results on Motion Deblurring with Gaussian Noise.}
    \label{fig:appendix_md}
    \vspace{-15pt} 
\end{figure}

\begin{figure}[H]
    \centering
    \includegraphics[width=1\linewidth]{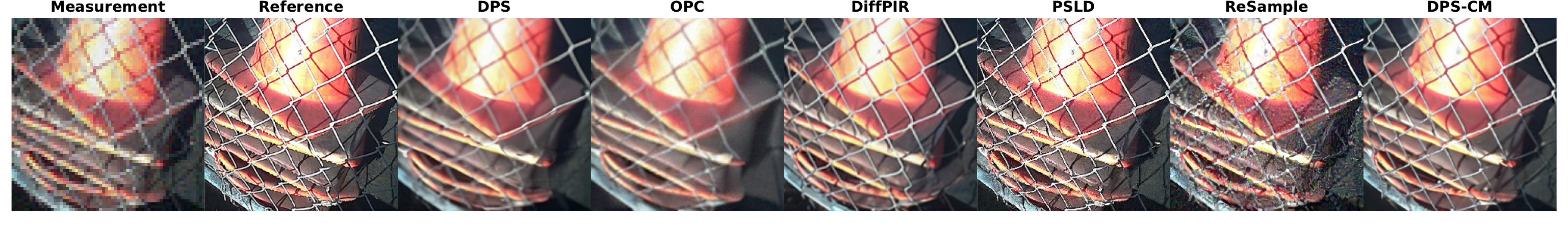}
    \caption{Additional Qualitative Results on super-resolution with Gaussian Noise.}
    \label{fig:appendix_sr}
    \vspace{-15pt}
\end{figure}

\begin{figure}[H] 
    \centering
    \includegraphics[width=1\linewidth]{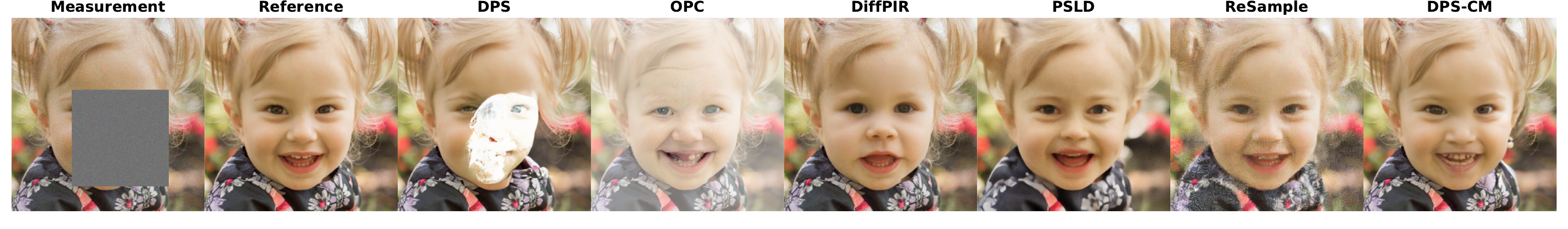}
    \caption{Additional Qualitative Results on Box Inpainting with Gaussian Noise.}
    \label{fig:appendix_box}
    \vspace{-15pt} 
\end{figure}

\begin{figure}[H] 
    \centering
    \includegraphics[width=1\linewidth]{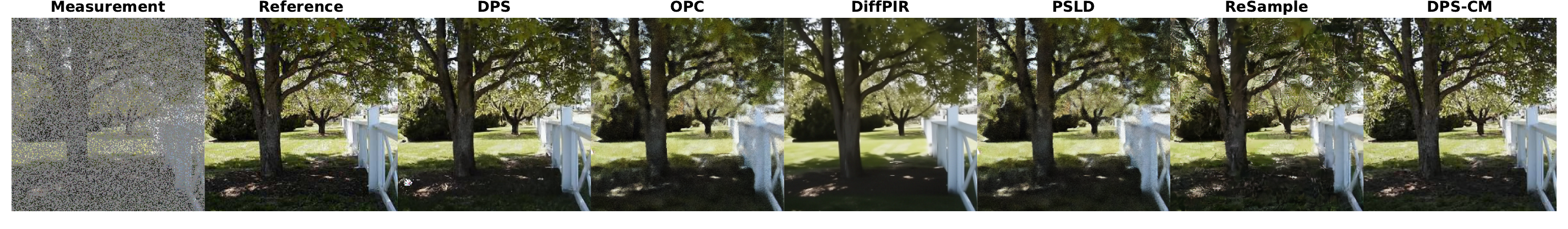}
    \caption{Additional Qualitative Results on Random Inpainting with Gaussian Noise.}
    \label{fig:appendix_random}
    \vspace{-15pt} 
\end{figure}

\begin{figure}[H] 
    \centering
    \includegraphics[width=0.7\linewidth]{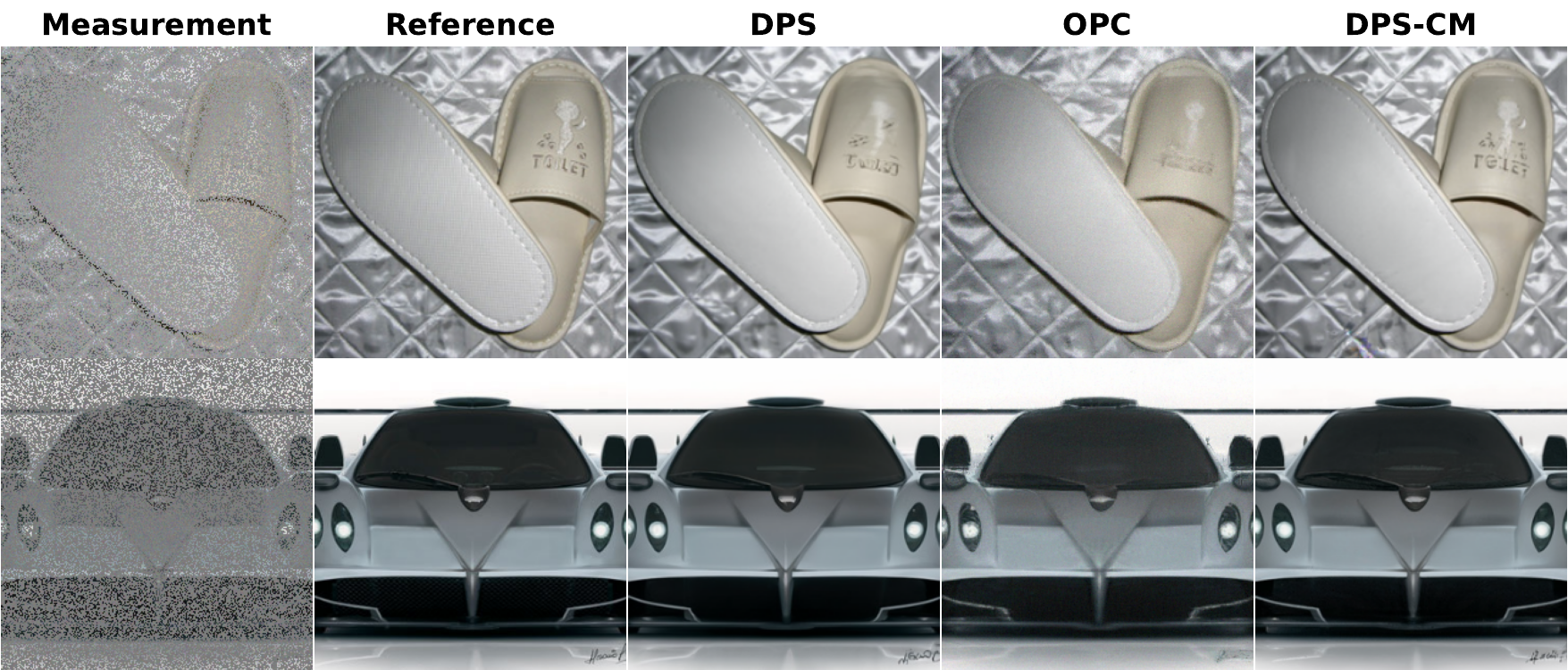}
    \caption{Additional Qualitative Results on Random Inpainting with Gaussian Noise.}
    \label{fig:ri}
    \vspace{-15pt} 
\end{figure}

\begin{figure}[H] 
    \centering
    \includegraphics[width=1\linewidth]{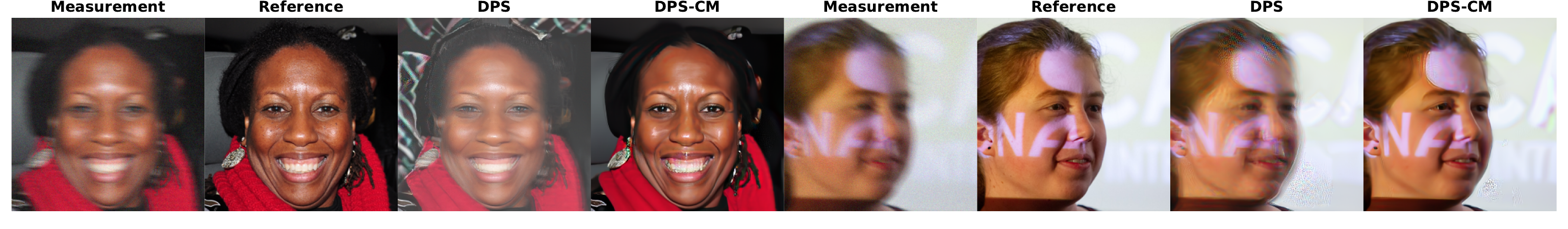}
    \caption{Additional Qualitative Results on Nonlinear Deblurring with Gaussian Noise.}
    \label{fig:appendix_nd}
    \vspace{-15pt} 
\end{figure}

\begin{figure}[H] 
    \centering
    \includegraphics[width=0.55\linewidth]{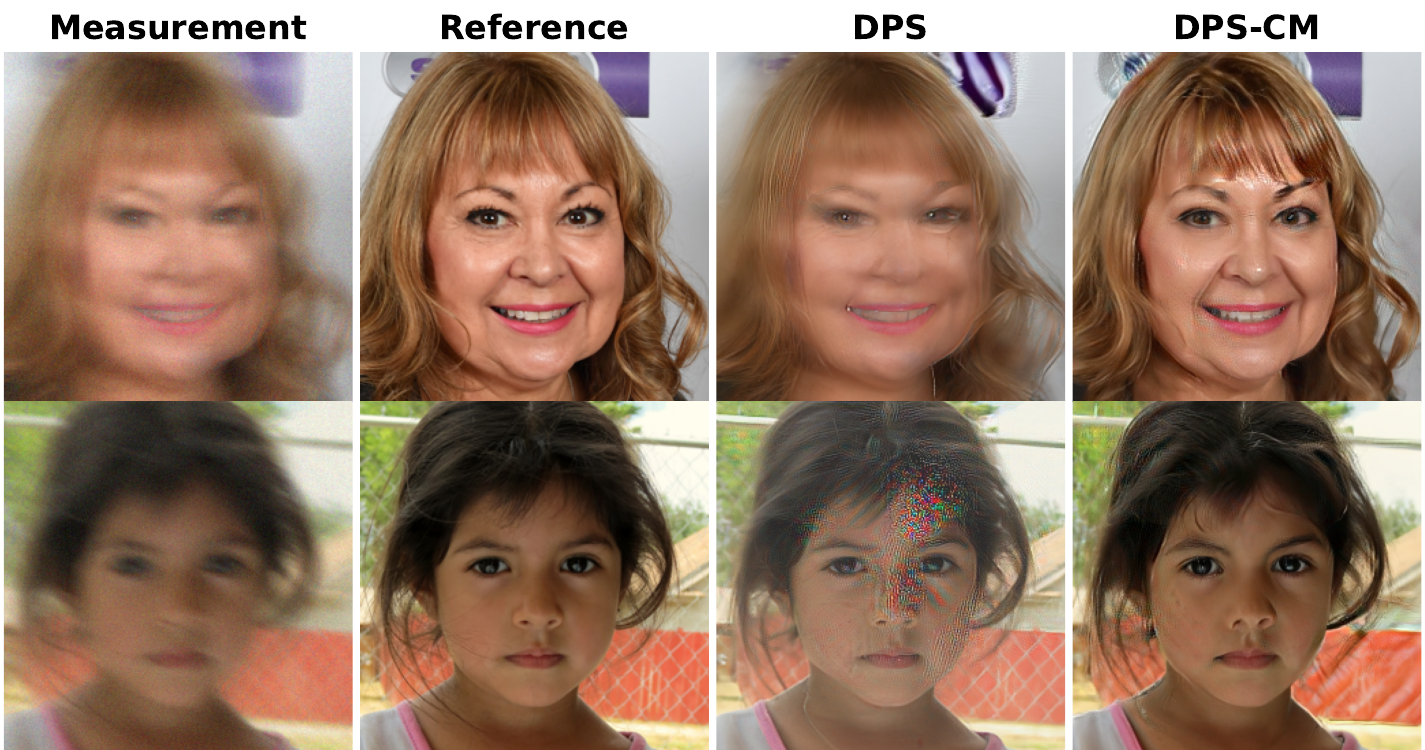}
    \caption{Additional Qualitative Results on Nonlinear Deblurring with Gaussian Noise.}
    \label{fig:nb}
    \vspace{-15pt} 
\end{figure}

\begin{figure}[H] 
    \centering
    \includegraphics[width=1\linewidth]{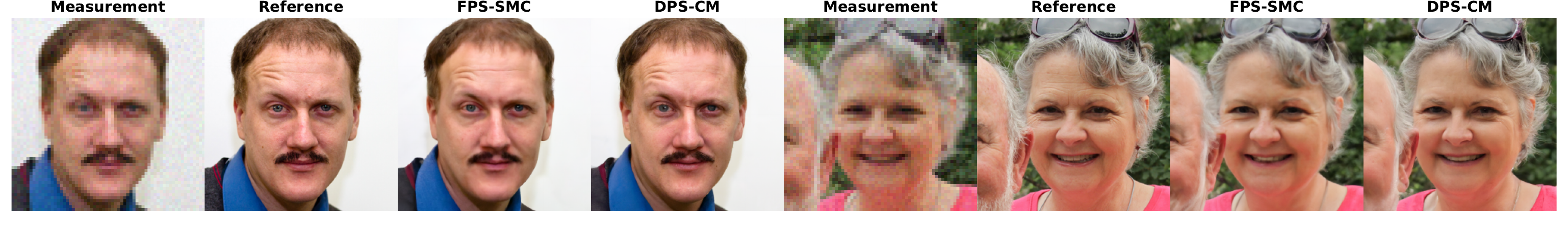}
    \caption{Additional Qualitative Results on Super-resolution with Gaussian noise compared with FPS-SMC ($M=10$).}
    \label{fig:appendix_sr_fps}
    \vspace{-15pt} 
\end{figure}

\begin{figure}[H] 
    \centering
    \includegraphics[width=0.7\linewidth]{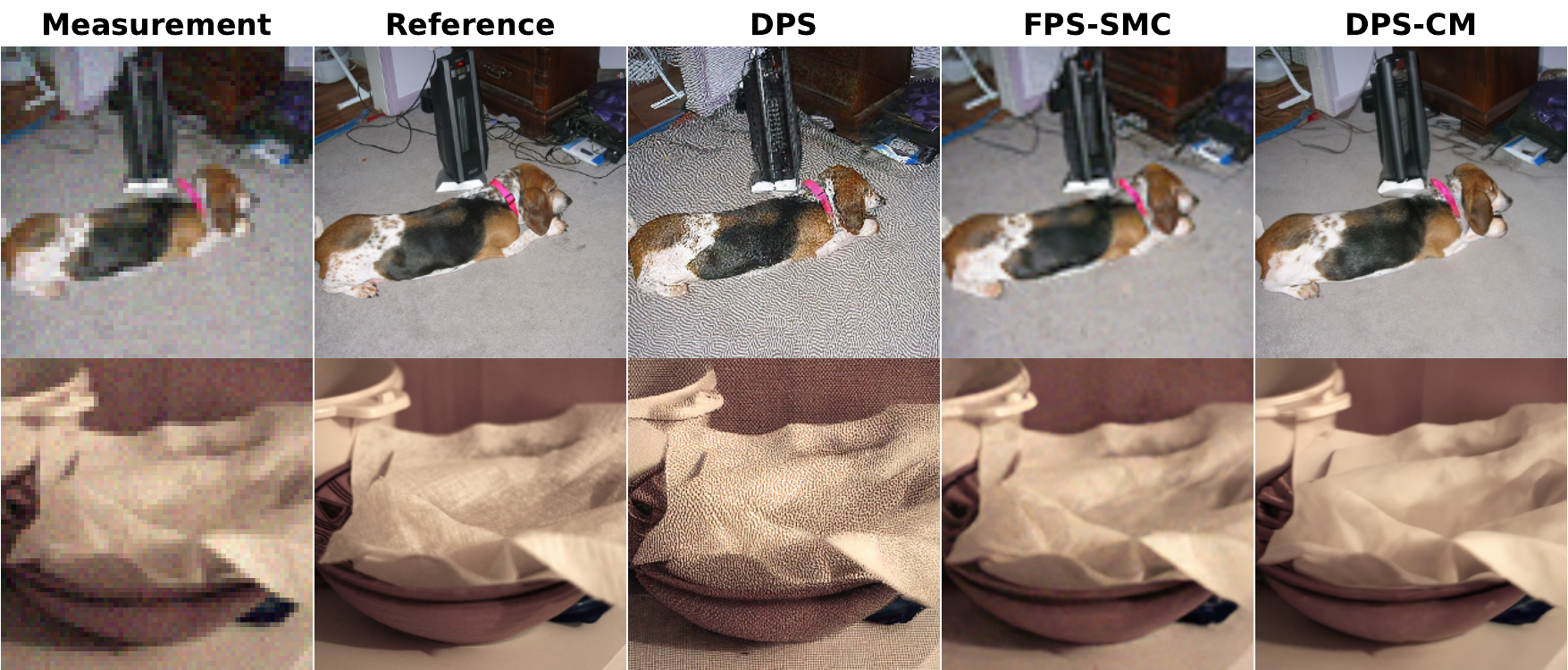}
    \caption{Additional Qualitative Results on Super-resolution with Gaussian Noise.}
    \label{fig:sr}
    \vspace{-15pt} 
\end{figure}

\begin{figure}[H]
    \centering
    \includegraphics[width=1\linewidth]{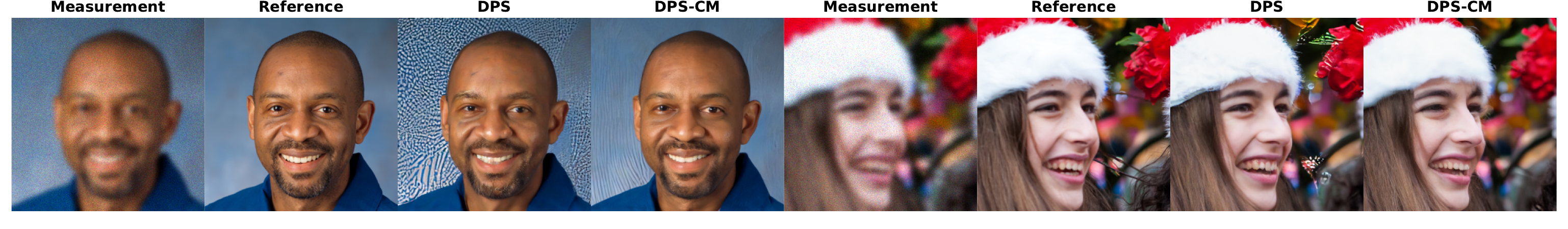}
    \caption{Additional Qualitative Results on Gaussian Deblurring with Poisson noise.}
    \label{fig:appendix_gd_poisson}
    \vspace{-15pt} 
\end{figure}

\begin{figure}[H]
    \centering
    \includegraphics[width=1\linewidth]{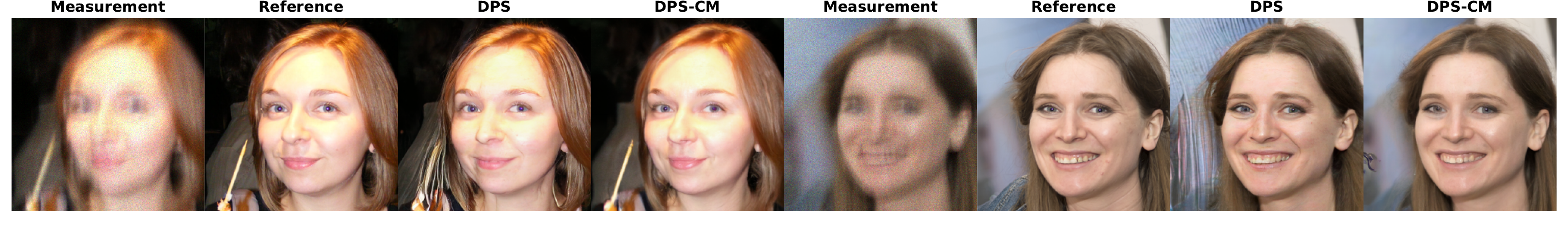}
    \caption{Additional Qualitative Results on Motion Deblurring with Poisson noise.}
    \label{fig:appendix_md_poisson}
    \vspace{-15pt} 
\end{figure}

\begin{figure}[H]
    \centering
    \includegraphics[width=1\linewidth]{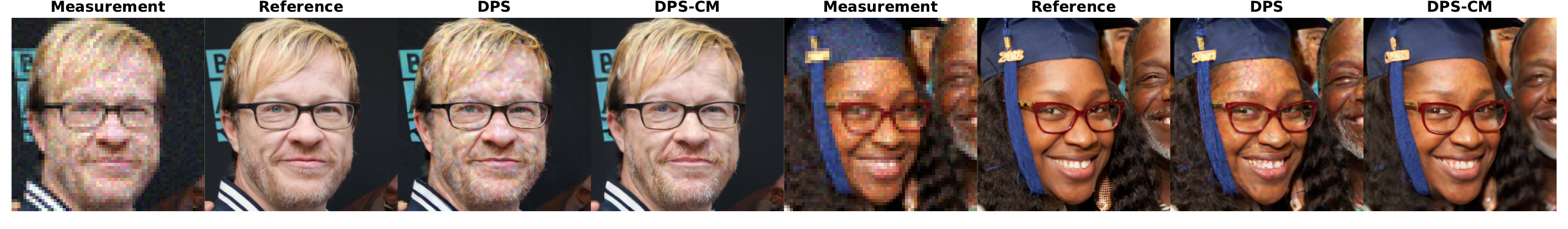}
    \caption{Additional Qualitative Results on Super-resolution with Poisson noise.}
    \label{fig:appendix_sr_poisson}
    \vspace{-15pt} 
\end{figure}

Additional visual examples are shown here to compare with baselines on different tasks with Gaussian noise ($\sigma=0.05$) from Fig.~\ref{fig:appendix_gd} to Fig.~\ref{fig:sr} and with Poisson noise ($\lambda=1.0$) from Fig.~\ref{fig:appendix_gd_poisson} to Fig.~\ref{fig:appendix_sr_poisson}. 

